\documentclass[twoside]{article}

\usepackage[preprint]{aistats2026}

\usepackage{graphicx}
\usepackage{amsmath}
\usepackage{amssymb}
\usepackage{amsfonts}
\usepackage{dsfont}
\usepackage[round]{natbib}
\usepackage{xurl}
\usepackage[disable]{todonotes}

\begin{document}

\runningtitle{Accelerating LLM Inference with Dynamic Early Exits}
\runningauthor{Valade}

\twocolumn[

\aistatstitle{Accelerating Large Language Model Inference with\\ Dynamic Self-Speculative Decoding}

\aistatsauthor{ Florian Valade }

\aistatsaddress{ Université Gustave Eiffel \\
Fujitsu}

]

\newcommand{\flo}[2][]{\todo[color=red!20,#1]{{\bf Flo:} #2}}
\newcommand{\moh}[2][]{\todo[color=cyan!20,#1]{{\bf Moh:} #2}}
\newcommand{\one}[1]{\mathds{1}_{\left\{#1\right\}}}

%\begin{abstract}
%This paper presents a modular approach to accelerate inference in large language models (LLMs) by adding early exit heads at intermediate transformer layers. Each head is trained in a self-supervised manner to mimic the main model's predictions, allowing computation to stop early when a calibrated confidence threshold is reached. We evaluate several confidence metrics and show that entropy provides the most reliable separation between correct and incorrect predictions. Experiments on the Pythia model suite (70M to 2.8B parameters) demonstrate that our method significantly reduces inference cost while maintaining accuracy across multiple benchmarks. We further adapt this approach to speculative decoding, introducing Dynamic Self-Speculative Decoding (DSSD), which achieves 1.66$\times$ higher token acceptance than manually-tuned LayerSkip baselines with minimal hyperparameter tuning.

%\end{abstract}

\section{Introduction}
\label{sec:introduction}

Large language models (LLMs) have become central to advancing capabilities in natural language processing (NLP), delivering remarkable performance across a range of tasks. The trend towards scaling up these models correlates strongly with improved performance, understanding, and generality. This relationship has been formalized through empirical scaling laws, which demonstrate that model performance improves predictably with increased model size, dataset size, and compute budget~\cite{kaplan_scaling_2020, hoffmann_training_2022}. However, the computational cost associated with these larger models is substantial, often necessitating the use of powerful server infrastructure~\cite{samsi_words_2023}. This not only limits local usability but also raises significant privacy concerns and requires considerable investment to scale in response to user demand. Solutions exist to reduce the computational demands of these models, but they often impact the model's performance by reducing its accuracy~\cite{zhu_survey_2023}.

Despite their effectiveness, these models often operate inefficiently. The nature of language itself contributes to this inefficiency; namely, not all tokens generated during the inference process contribute equally to the overall meaning or require the same level of computational resources. Some tokens are inherently simpler and can be predicted with high confidence early in the computation process, while others, contributing more significantly to the context or meaning, may require deeper processing.

In response to these challenges, we develop a method that can be easily integrated into existing pre-trained models to enhance their inference speed without extensive retraining. Our solution focuses on the strategic placement of early exit "heads"~\cite{Scardapane2020Why,teerapittayanon2017branchynet} within the transformer layers of an LLM. These heads terminate the inference process when a calibrated confidence threshold is met, based on the complexity and predictability of the token being processed.

Our contributions are twofold:
\begin{enumerate}
    \item We provide a detailed experimental study and modular framework for training and deploying early exit heads on top of LLMs. We analyze multiple training strategies, confidence metrics, and demonstrate scalability across model sizes from 70M to 2.8B parameters on the Pythia suite.
    \item We adapt our early exit mechanism to speculative decoding, introducing \textbf{Dynamic Self-Speculative Decoding (DSSD)}, which achieves \textbf{1.66$\times$ higher token acceptance rates} than manually-tuned LayerSkip baselines while requiring minimal hyperparameter tuning—only a single accuracy threshold $\epsilon$.
\end{enumerate}

\section{Related Work}
\label{sec:relatedwork}

Transformers~\cite{vaswani_attention_2017} scaled into today’s LLM families such as BERT, GPT-3, PaLM, LaMDA, LLaMA and OPT~\cite{devlin_bert_2019,brown_language_2020,chowdhery_palm_2022,thoppilan_lamda_2022,touvron_llama_2023,zhang_opt_2022}, powering vision~\cite{dosovitskiy_image_2020}, speech~\cite{radford_robust_2023} and multimodal models.  Their billion-parameter footprints, however, make every token generation costly.  Static compression—quantisation, pruning and distillation~\cite{shen_q-bert_2019,sun_mobilebert_2020,zeroquant_yao_2022,fan_reducing_2019,sun_patient_2019,bai_binarybert_2021}—slashes model size but still expends identical compute on easy and hard inputs.  Early-exit methods address this imbalance by attaching lightweight classifiers to intermediate layers and halting computation once a confidence criterion is met.

In computer vision, BranchyNet~\cite{teerapittayanon2017branchynet} and MSDNet~\cite{huang2018multiscale} or EERO methodology~\cite{valade2024eeroearlyexitreject} established the two key components still used today: deeply supervised branches and an entropy-based exit rule.  Transferring the idea to Transformers, DeeBERT and \emph{The Right Tool} calibrated soft-max confidence to save up to $5\times$ latency with negligible loss~\cite{xin_deebert_2020,schwartz_righttool_2020}.  FastBERT distilled the final head into earlier exits~\cite{liu2020fastbert}, PABEE demanded $k$ consecutive agreeing predictions instead of a threshold~\cite{zhou_bert_2020}, and BERxiT learned an explicit when-to-exit module while alternating fine-tune schedules~\cite{jitkarl_berxit_2021}.  Skip/SmartBERT added trainable gates that may bypass whole layers, combining skipping with exiting for $2$–$4\times$ cheaper inference~\cite{li_skipbert_2022,chen_smartbert_2023}.

Early exit for sequence-to-sequence generation is newer.  Depth-Adaptive Transformers first applied exits to neural MT~\cite{elbayad_depthadaptive_2020}; FREE and DEED refined token-level uncertainty estimates~\cite{wang_free_2023,yang_deed_2024}.  Recent evidence shows exits remain effective inside 13 B-parameter LLaMA-2 and GPT-J~\cite{liu_earlyexitllama_2024}, suggesting scalability to modern LLMs.

Speculative decoding~\cite{leviathan2023fast,chen2023accelerating} accelerates LLM inference by using a smaller draft model to generate candidate tokens, which are then verified in parallel by the target model. LayerSkip~\cite{elhoushi2024layerskip} combines this with early exits, using intermediate layers as draft models. However, LayerSkip requires exhaustive hyperparameter search over both head selection and speculation length. Our DSSD method addresses this limitation by adaptively selecting exit layers based on calibrated confidence.

Our work inherits the plug-and-play nature of DeeBERT-style branches, borrows the self-supervised signal of FastBERT, and scales token-level exits to different architecture sizes.  We train exit heads without extra data—using the model's own probabilities—then calibrate a single threshold on a held-out set à la conformal prediction~\cite{Vovk99IntroCP}.  Unlike layer-skipping approaches, the backbone weights remain frozen, guaranteeing monotonic accuracy with deeper computation.

\section{Methodology}
\label{sec:methodology}

This section details our methodology for integrating and utilizing early exits within large language models (LLMs) to enhance computational efficiency during inference. The approach is designed to be generalizable and, while we demonstrate its application using the Pythia suite, it is applicable to any multi-layered transformer model. This adaptability ensures that our methodology can be leveraged across a broad spectrum of modern LLMs, enhancing their usability without requiring significant modifications to their underlying architectures.

\subsection{Definitions and Notation}

We introduce here the main notations used throughout the paper.

\paragraph{Vocabulary.}
Let \(\mathcal{V}\) denote the vocabulary, a finite set of tokens:
\[
    \mathcal{V} = \{v_1, v_2, \dots, v_{|\mathcal{V}|}\},
\]
where $|\cdot|$ denotes cardinality of sets.
\paragraph{Dataset.}
We consider a dataset \(\mathcal{D}\) of \(N\) examples, where each example is a sequence of tokens:
\[
    \mathcal{D} = \{\mathbf{x}^{(i)}\}_{i=1}^N, \quad \mathbf{x}^{(i)} = (x^{(i)}_1, \dots, x^{(i)}_{L_i}), \quad x^{(i)}_j \in \mathcal{V}.
\]
A subset \(\mathcal{D}_{\text{cal}} \subset \mathcal{D}\) is reserved as a calibration set, the rest being used for training.

Each sequence \(\mathbf{x}^{(i)}\) can have a variable length \(L_i\). The total number of tokens in the dataset is \(M = \sum_{i=1}^N L_i\), with \(M > N\) in general. Similarly, the calibration set \(\mathcal{D}_{\text{cal}}\) contains \(N_{\text{cal}}\) examples and a total of \(M_{\text{cal}}\) tokens.

\paragraph{Language model.}
A large language model (LLM) is a function \(f_\theta\) parameterized by \(\theta\), mapping a sequence of tokens to a sequence of probability distributions over the vocabulary:
\[
    f_\theta: (x_1, \dots, x_L) \mapsto (\mathbf{p}_1, \dots, \mathbf{p}_L), \quad \mathbf{p}_t \in \Delta^{|\mathcal{V}|},
\]
where \(\Delta^{|\mathcal{V}|}\) is the probability simplex over \(\mathcal{V}\).

\paragraph{Objective.}
Given an input sequence \(\mathbf{x} = (x_1, \dots, x_L) \in \mathcal{V}^L  \), the target sequence is \(\mathbf{y} = (y_1, \dots, y_L)\in \mathcal{V}^L  \) with \(y_{t} = x_{t+1}\) (i.e., the next-token prediction task).

\paragraph{Early exit heads.}
We introduce \(K\) early exit heads, each defined as a function \(h_k\) (with its own parameters, but sharing the backbone with \(f_\theta\)) that maps a sequence of $L$ tokens to a sequence of probability distributions:
\[
    h_k: (x_1, \dots, x_L) \mapsto (\mathbf{p}_{k,1}, \dots, \mathbf{p}_{k,L}), \quad \mathbf{p}_{k,t} \in \Delta^{|\mathcal{V}|}.
\]
All early exit heads share the same backbone, so most of their parameters are shared with the main model.

\paragraph{Confidence metric.}
A confidence metric is a function \(c: \Delta^{|\mathcal{V}|} \to \mathbb{R}\) that assigns a real-valued confidence score to a probability vector (e.g., \(c(\mathbf{p}) = \max_j p_j\)).

\paragraph{Accuracy threshold $\epsilon$.}
The parameter $\epsilon \in [0,1]$ controls the minimal desired accuracy for early exit decisions. For each early exit head, a prediction is only output if its confidence metric exceeds a calibrated threshold corresponding to at least $\epsilon$ empirical accuracy on a held-out calibration set. Lowering $\epsilon$ increases speedup at the cost of potential accuracy loss, while higher $\epsilon$ enforces stricter correctness guarantees.

The above definitions give a precise, token-level view of the model outputs and their interaction with early-exit logic; they will serve as the foundation for the training objectives, calibration techniques, and inference algorithms described in the following sections.

\subsection{Implementation Details}
\label{subsec:implementationtraining}

To enhance the inference efficiency of large language models, we incorporate early exit "heads" into a pre-existing model, in this instance, the Pythia suite~\cite{biderman2023pythiasuiteanalyzinglarge}. These heads are implemented at regular intervals along the network. Structurally, each head is a simple multi-layer perceptron (MLP) identical to the final classification head of the model.
Each of these head takes as input hidden features from a transformer block inside the model.
 
\paragraph{Placement of early–exit heads.}
Let the transformer backbone consist of \(L\) stacked blocks, indexed
\(
    \mathcal{I}
    =\{1,2,\dots,L\},
\)
and let \(K\) denote the desired number of early–exit heads.  
To distribute the heads uniformly while keeping the final classification
layer untouched, we attach head \(k\in\{1,\dots,K\}\) to the block whose
index is
\begin{equation}
    \label{eq:indexblock}
    \ell_k
    \;=\;
    \Bigl\lfloor
        \frac{k}{K+1}\,L
    \Bigr\rfloor,
    \qquad
    k=1,\dots,K.
\end{equation}
Equation~\eqref{eq:indexblock} simply divides the layer indices into \(K+1\) equal
segments and selects the endpoint of each segment (rounded down to the
nearest integer) as a branch location.  
In our experiments we set \(K=4\); hence the heads are inserted at
\(
    \ell_k\in\bigl\{\lfloor\tfrac{L}{5}\rfloor,\;
                   \lfloor\tfrac{2L}{5}\rfloor,\;
                   \lfloor\tfrac{3L}{5}\rfloor,\;
                   \lfloor\tfrac{4L}{5}\rfloor
             \bigr\}.
\)

For the implementation of these heads, we experimented with two initialization strategies: initializing the heads from scratch and copying the final classification head in order to fine-tune it. The difference between the two are analyzed in Section~\ref{sec:experiments}.
With the architectural choices established, we next describe the data and objectives used to train the early exit heads.

\subsection{Training data and objectives}

\paragraph{Corpus.}
All auxiliary heads are trained on \textsc{MiniPile}~\cite{kaddour2023minipile}, a 6-GB stratified
subset of the original 825-GB \textsc{ThePile-deduplicated}~\cite{gao2020pile} dataset that was
employed for pre-training the \textsc{Pythia} backbone
\cite{biderman2023pythiasuiteanalyzinglarge}.  Using the same data
distribution avoids the distribution-shift issues that often appear
when auxiliary classifiers are fitted post-hoc.

\paragraph{Losses.}
For the whole predicted sequence \(\mathbf{\hat{y}}\), we can compute different losses depending on the objectives. We define the cross-entropy loss and the Kullback–Leibler divergence between the predicted probability distribution \(\hat{y}\) and the ground-truth labels \(y\) as follows:
\begin{equation*}
\mathcal{L}_{\text{CE}}(\mathbf{y}, \mathbf{\hat{y}}) = -\frac{1}{L}\sum_{t=1}^{L} \sum_{v \in \mathcal{V}} \mathbf{1}_{\{\mathbf{y}^{(t)} = v\}} \log(\mathbf{\hat{y}}^{(t)}_v),
\end{equation*}
\begin{equation*}
    \mathcal{L}_{\text{KL}}(\mathbf{\hat{y}} || \mathbf{y}) = \frac{1}{L}\sum_{t=1}^{L} \sum_{v \in \mathcal{V}}\mathbf{\hat{y}}_v \log{\frac{\mathbf{\hat{y}}_v}{\mathbf{y}_v}}.
\end{equation*}
In the context of self-supervised training, we consider the Kullback-Leibler divergence where the target $\textbf{y}$ is replaced by the output of the main model $f_\theta$. Thus, the loss $\mathcal{L}_{\text{KL}}$ becomes:
\begin{equation*}
    \mathcal{L}_{\text{KL}}(\mathbf{\hat{y}} || f_\theta) = \frac{1}{L}\sum_{t=1}^{L} \sum_{v \in \mathcal{V}}\mathbf{\hat{y}}_v \log{\frac{\mathbf{\hat{y}}_v}{f_\theta(v)}}
\end{equation*}
The cross-entropy loss \(\mathcal{L}_{\text{CE}}\) is used when considering a supervised training objective. It matches how the main model is trained and can be used to train the heads as well.
The Kullback–Leibler divergence \(\mathcal{L}_{\text{KL}}\) is used when considering a self-supervised training objective. It encourages the early exit heads to mimic the full probability distribution of the main model, which can be useful for improving the performance of the early exits.
From this, we consider three training objectives obtained from the above building
blocks:

\begin{itemize}
    \item \textbf{Supervised} (\(\mathcal{L}_{\text{sup}}\)).
          Purely next-token cross-entropy against ground-truth labels, that is,
          \(\mathcal{L}_{\text{sup}} = \mathcal{L}_{\text{CE}}\);
    \item \textbf{Self-supervised} (\(\mathcal{L}_{\text{self}}\)).
          KL divergence that encourages each head to mimic the teacher’s
          full probability distribution, that is,
          \(\mathcal{L}_{\text{self}} = \mathcal{L}_{\text{KL}}\);
    \item \textbf{Hybrid} (\(\mathcal{L}_{\text{hyb}}\)).
          The sum of the two losses above, weighted by coefficients \(\alpha\in(0,1)\), that is,
          \(\mathcal{L}_{\text{hyb}} = \alpha\,\mathcal{L}_{\text{CE}} + (1 - \alpha) \,\mathcal{L}_{\text{KL}}\).
\end{itemize}
As reported in Section~\ref{sec:experiments}, the self-supervised
objective (\(\mathcal{L}_{\text{self}}\)) yields the most faithful
approximation of the teacher’s behavior while preserving calibration,
and therefore constitutes our default choice for all subsequent
experiments.

\subsection{Calibration and Inference}
\label{subsec:calibrationinference}

After training the early exit heads, the next crucial step involves calibrating and using these heads during model inference. This process is divided into two main stages: calibration of the confidence thresholds and the application of these thresholds during inference.

\subsubsection{Calibration of Confidence Thresholds}
\label{subsubsec:calibration}
\begin{figure}[ht]
    \centering
    \includegraphics[width=0.48\textwidth]{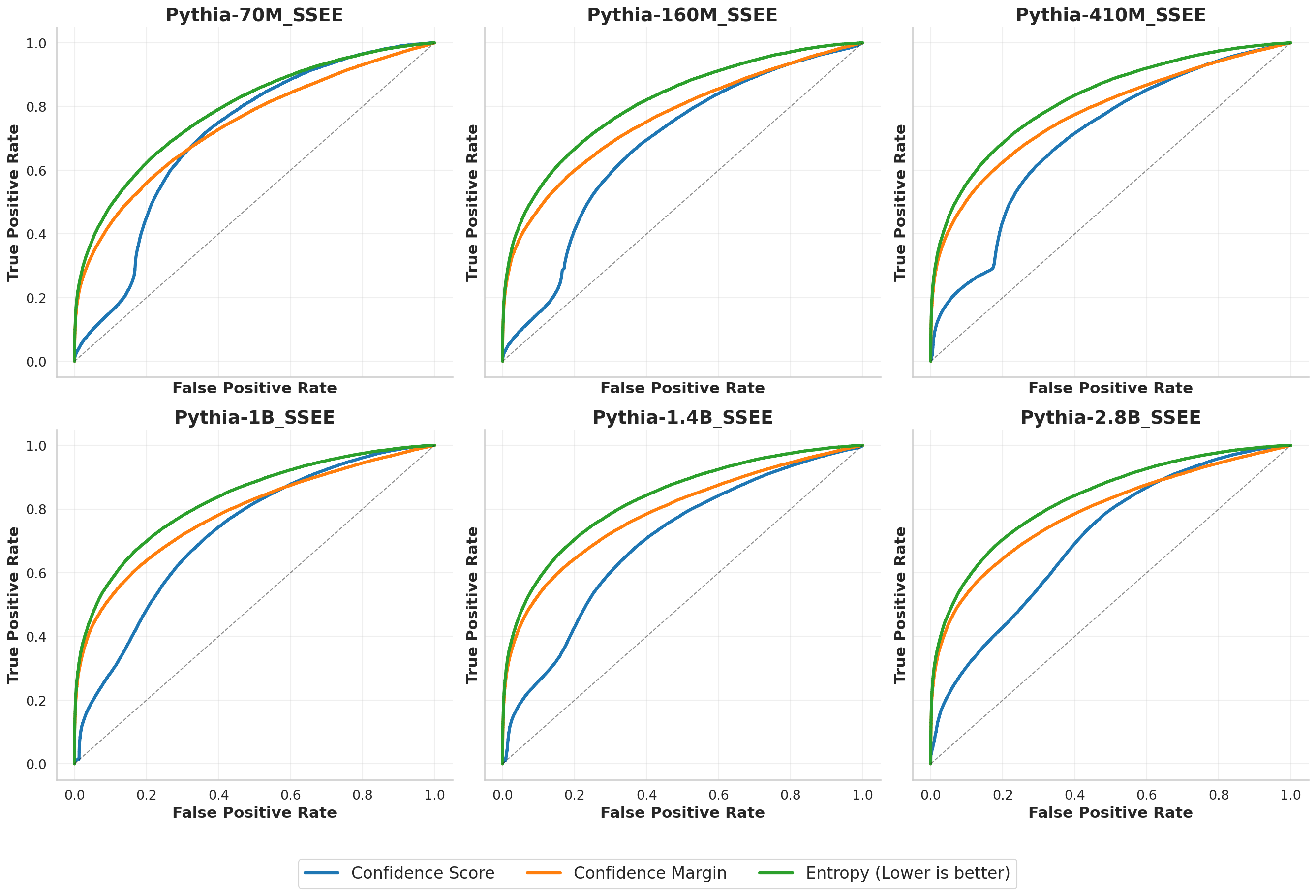}
    \caption{ROC curves for three confidence metrics across six Pythia models (70M to 2.8B). Entropy consistently achieves the highest AUC.}
    \label{fig:aggregate_roc_curves}
\end{figure}

In our approach, we evaluated three different confidence metrics for calibrating early exit thresholds: (1) maximum probability, (2) entropy of the predicted distribution, and (3) the difference between the top two probabilities ("breaking ties"). To determine which metric best separates correct from incorrect predictions, we analyzed their ability to discriminate between right and wrong outputs using ROC (Receiver Operating Characteristic) curves (Figure~\ref{fig:aggregate_roc_curves}).

After training, we computed the ROC curves for each metric by plotting the true positive rate against the false positive rate as the threshold varies, using the predictions from the early exit heads. This analysis was performed across six Pythia models, ranging from 70M to 2.8B parameters. The results are summarized in Figure~\ref{fig:aggregate_roc_curves}, which displays the ROC curves for all metrics and models.

Our findings consistently show that entropy outperforms the other metrics in every case, achieving a higher area under the curve (AUC) regardless of model size. This indicates that entropy provides a more reliable separation between correct and incorrect predictions, making it the most effective metric for threshold calibration in our early exit framework. Furthermore, we observe that the AUC of the entropy metric tends to increase with model size, suggesting that early exit mechanisms become even more effective as the underlying model grows larger.

To calibrate the confidence thresholds, we perform a full epoch over our calibration dataset. For each example in the calibration set and for each early exit head, we compute the head output for every token in the sequence, then apply the confidence metric $c$ to each token output, and store all these scores in a global vector $\mathbf{c}_k \in \mathbb{R}^{M_{\text{cal}}}$, where $M_{\text{cal}}$ is the total number of tokens in the calibration dataset.

Correspondingly, let \( \mathbf{t}_k \) be a binary vector of length \(M_{\text{cal}}\) where each element corresponds to the correctness of the prediction associated with the respective element in \( \mathbf{c}_k \). Specifically, \( t_{k,j} \) is 1 if the \( j^{th}\) prediction at head \( k \) matches the \( j^{th} \) prediction of the underlying model \(f_\theta(\mathbf{x})\), otherwise 0:
\begin{equation}
\label{eq:accuracy_calibrationSet}
    \begin{aligned}
    t_{k,j} &= \one{\arg\max \mathbf{p}_{k,j} = \arg\max \mathbf{p}_{\theta,j}} \\
    &= \left\{    
    \begin{array}{ll}
    1 & \text{if } \arg\max(\mathbf{p}_{k,j}) = \arg\max(\mathbf{p}_{\theta,j}) \\
    0 & \text{otherwise}
    \end{array}
    \right.
    \end{aligned}
\end{equation} 

To determine the confidence threshold for each early exit head, we utilize the calibration set to empirically estimate the relationship between the confidence metric and prediction correctness. Specifically, for each head \(k\), we sort the calibration metric values in ascending order. Given a user-specified confidence level \(\epsilon \in [0,1]\), which represents the minimum desired proportion of correct predictions above the threshold, we identify the smallest metric value such that the proportion of correct predictions among all samples with higher (or equal) metric values is at least \(\epsilon\). This procedure ensures that, during inference, predictions made with a confidence metric exceeding the threshold are correct with probability at least \(\epsilon\), thereby providing a principled trade-off between computational efficiency and predictive accuracy. Formally, the threshold for each head is defined as follows.

Recall that $\mathbf{c}_k$ and $\mathbf{t}_k$ denote, respectively, the vectors of confidence metric values and correctness indicators for head $k$, as defined in equation~\eqref{eq:accuracy_calibrationSet}. To proceed, we jointly sort these vectors in ascending order according to the values in $\mathbf{c}_k$, resulting in ordered sequences $\mathbf{c}_k = (c_{k,1}, c_{k,2}, \ldots, c_{k,M_{\text{cal}}})$ and $\mathbf{t}_k = (t_{k,1}, t_{k,2}, \ldots, t_{k,M_{\text{cal}}})$ such that $c_{k,1} \leq c_{k,2} \leq \ldots \leq c_{k,M_{\text{cal}}}$. 

We then define $\hat{j}$ as the smallest index for which the proportion of correct predictions among all samples with confidence at least $c_{k,\hat{j}}$ meets or exceeds the target confidence level $\epsilon$:
\begin{equation*}
    \frac{\sum_{i=\hat{j}}^{M_{\text{cal}}} t_{k,i}}{M_{\text{cal}} - \hat{j} + 1} \geq \epsilon.
\end{equation*}
Then, let the threshold \(\tau_k\) be defined as:
\begin{equation*}
    \tau_k = c_{k,\hat{j}}.
\end{equation*}
In other words, \(\tau_k\) is the value of the confidence metric at the index \(\hat{j}\), where \(\hat{j}\) is the smallest index such that the proportion of correct prediction in the remaining samples is at least \(\epsilon\).
%TODO in other word when reducing epsilon we augment speeup
%Also add epsilon in contribution and definition
\subsubsection{Inference Process}
\label{subsubsec:inferenceprocess}

After establishing the confidence thresholds for each early exit head through the calibration process, the model is then ready to utilize these thresholds during the inference phase to efficiently process new inputs.

During inference, each input \(\mathbf{x}\) is sequentially processed through the model's layers, with the possibility of early termination at any of the early exit heads \(h_k\). For each head \(k \in \{1, 2, \ldots, K\}\), we compute:
\begin{equation*}
\begin{split}
    \mathbf{p}_k &= h_k(\mathbf{x}), \\
    c_k &= c(\mathbf{p}_k).
\end{split}
\end{equation*}
We then return \(\mathbf{p}_k\) as output from the model if:
\begin{equation*}
    c_k \geq \tau_k.
\end{equation*}
If no head satisfies this inequality, we return \(\mathbf{p}_{\theta}\).

This calibrated and threshold-driven early exit mechanism allows the model to balance efficiency with accuracy, ensuring that resource-intensive computations are only performed when necessary.

\section{Experiments and Results}
\label{sec:experiments}

We organize our experimental evaluation into two complementary parts\footnote{All computations are run on a server with an Intel(R) Xeon(R) Gold 5120 CPU and a Tesla V100 GPU with 32GB of Vram and 64GB of RAM. Code used in experiments to train and evaluate can be found under : https://anonymous.4open.science/r/BranchyLLM-B870}. First, we evaluate calibrated early exits on standard inference tasks using the Pythia suite (70M--2.8B parameters). We describe the training process for early exit heads and analyze how accuracy-speedup trade-offs scale across model sizes on benchmark tasks. Second, we extend our method to speculative decoding, introducing Dynamic Self-Speculative Decoding (DSSD), which eliminates the need for manual hyperparameter tuning. Throughout our experiments, we use Pythia as a representative example that allows us to scale according to hardware capabilities, though our approach is theoretically applicable to any language model architecture.

\subsection{Training}

In the training phase, we systematically compared the three loss functions described in Section~\ref{subsec:implementationtraining}: supervised (cross-entropy), self-supervised (KL divergence to the main model's output), and a hybrid of both. Our primary goal was to select a training objective that enables early exit heads to best approximate the main model's predictions while also providing meaningful uncertainty estimates.

We found that the self-supervised loss, which encourages each early exit head to mimic the output distribution of the main model, consistently led to the best alignment with the main model's predictions. This loss not only matches the output probabilities but also preserves the calibration properties necessary for reliable early exits.

Additionally, we experimented with two initialization strategies for the early exit heads: (1) copying the weights from the main model's final classification head (lm\_head), and (2) random initialization. When the early exit heads are initialized by copying the lm\_head, they start with a lower loss and initially perform better. However, as training progresses, the randomly initialized heads quickly overtake the copied ones in both loss and performance.

A key observation is that heads copied from the lm\_head tend to be overconfident, even when making incorrect predictions. This results in low entropy outputs and a lack of meaningful uncertainty, which is detrimental for early exit decisions. In contrast, randomly initialized heads, when trained with the self-supervised loss, develop a better notion of uncertainty, producing higher entropy outputs when unsure. This property is crucial for effective early exit mechanisms, as it allows the confidence metric to reliably distinguish between correct and incorrect predictions.

For our training experiments, we launch training with the self-supervised loss described above on all Pythia models from 70M to 2.8B parameters. We use a learning rate of $5\times10^{-5}$ and train on 500,000 examples sampled from the MiniPile dataset. Four early exit heads are added at the locations described in the Methodology section.

\subsection{Inference}
\label{subsec:inference}

\begin{figure}[tbp!]
    \centering
    \includegraphics[width=0.48\textwidth]{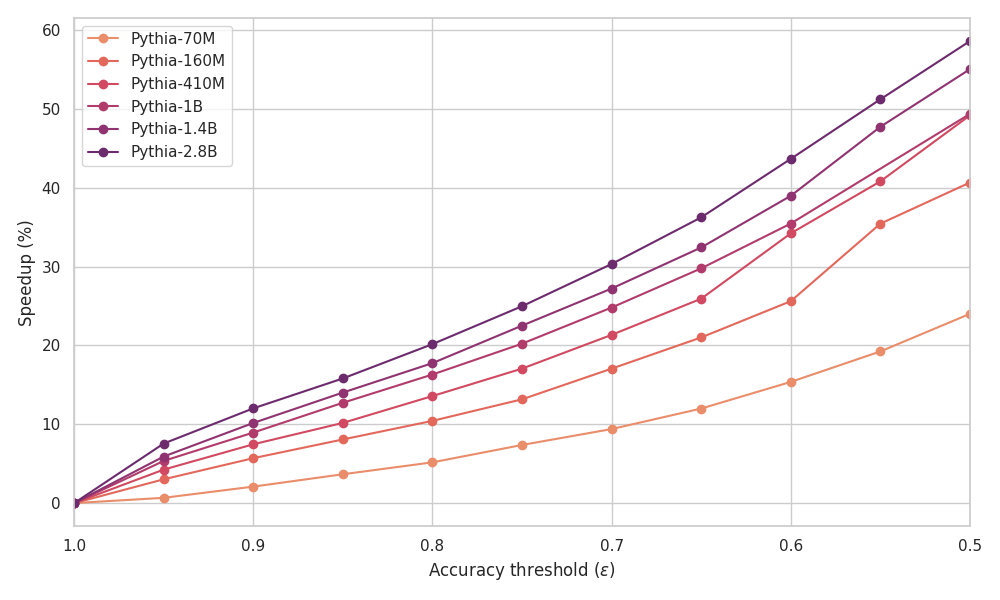}
    \caption{Speedup vs. confidence threshold $\epsilon$ for different Pythia model sizes.}
    \label{fig:all_models_speedup_vs_threshold}
\end{figure}

For the inference evaluation, each model is evaluated with different values of the confidence threshold parameter $\epsilon$, ranging from 0.5 to 1.0 in increments of 0.05. This systematic sweep allows us to analyze the trade-off between computational savings and predictive accuracy.

We follow the same set of benchmarks as the original Pythia paper: WSC, Winogrande, SciQ, PIQA, LogiQA, LAMBADA\_OpenAI, ARC Easy, and ARC Challenge. For each benchmark and each value of $\epsilon$, we report both the benchmark score and the mean output layer, translated to a speedup percentage.

\begin{table}[htbp!]
\footnotesize
\centering
\caption{Benchmarks results for Pythia-2.8B. Each line corresponds to a value of $\epsilon$ ranging from 0.5 to 1.0 in steps of 0.05 (top to bottom), with the bottom line being the baseline ($\epsilon=1$). More benchmarks available in the appendix.}
\begin{tabular}{lccc}
\hline
$\epsilon$ & Speedup & winogrande & piqa \\
\hline
0.50 & {\color[RGB]{0,221,0} 58.6\%} & {\color[RGB]{192,0,0} 0.541} $\pm$ {\color[RGB]{0,0,0} 0.014} & {\color[RGB]{255,0,0} 0.615} $\pm$ {\color[RGB]{0,0,0} 0.011} \\
0.55 & {\color[RGB]{0,205,0} 51.2\%} & {\color[RGB]{0,0,0} 0.564} $\pm$ {\color[RGB]{0,0,0} 0.014} & {\color[RGB]{255,0,0} 0.635} $\pm$ {\color[RGB]{0,0,0} 0.011} \\
0.60 & {\color[RGB]{0,190,0} 43.7\%} & {\color[RGB]{0,0,0} 0.565} $\pm$ {\color[RGB]{0,0,0} 0.014} & {\color[RGB]{255,0,0} 0.643} $\pm$ {\color[RGB]{0,0,0} 0.011} \\
0.65 & {\color[RGB]{0,174,0} 36.2\%} & {\color[RGB]{0,0,0} 0.565} $\pm$ {\color[RGB]{0,0,0} 0.014} & {\color[RGB]{255,0,0} 0.668} $\pm$ {\color[RGB]{0,0,0} 0.011} \\
0.70 & {\color[RGB]{0,162,0} 30.3\%} & {\color[RGB]{0,144,0} 0.586} $\pm$ {\color[RGB]{0,0,0} 0.014} & {\color[RGB]{255,0,0} 0.681} $\pm$ {\color[RGB]{0,0,0} 0.011} \\
0.75 & {\color[RGB]{0,151,0} 25.0\%} & {\color[RGB]{0,0,0} 0.579} $\pm$ {\color[RGB]{0,0,0} 0.014} & {\color[RGB]{255,0,0} 0.690} $\pm$ {\color[RGB]{0,0,0} 0.011} \\
0.80 & {\color[RGB]{0,141,0} 20.2\%} & {\color[RGB]{0,0,0} 0.574} $\pm$ {\color[RGB]{0,0,0} 0.014} & {\color[RGB]{207,0,0} 0.707} $\pm$ {\color[RGB]{0,0,0} 0.011} \\
0.85 & {\color[RGB]{0,132,0} 15.8\%} & {\color[RGB]{0,0,0} 0.579} $\pm$ {\color[RGB]{0,0,0} 0.014} & {\color[RGB]{170,0,0} 0.719} $\pm$ {\color[RGB]{0,0,0} 0.010} \\
0.90 & {\color[RGB]{0,124,0} 12.0\%} & {\color[RGB]{0,0,0} 0.571} $\pm$ {\color[RGB]{0,0,0} 0.014} & {\color[RGB]{135,0,0} 0.730} $\pm$ {\color[RGB]{0,0,0} 0.010} \\
0.95 & {\color[RGB]{0,115,0} 7.6\%} & {\color[RGB]{0,0,0} 0.572} $\pm$ {\color[RGB]{0,0,0} 0.014} & {\color[RGB]{0,0,0} 0.733} $\pm$ {\color[RGB]{0,0,0} 0.010} \\
1.00 & {\color[RGB]{0,0,0} 0.0\%} & {\color[RGB]{0,0,0} 0.571} $\pm$ {\color[RGB]{0,0,0} 0.014} & {\color[RGB]{0,0,0} 0.742} $\pm$ {\color[RGB]{0,0,0} 0.010} \\
\hline
\end{tabular}
\label{tab:pythia2.8b_part1}
\end{table}

The results of the benchmark on Pythia-2.8B are shown in Table~\ref{tab:pythia2.8b_part1}. The results for the other model sizes are provided in the appendix.

Our results show that for some benchmarks, such as Winogrande and WSC, the reduction in computation has little to no effect on accuracy, even with aggressive early exiting. However, some benchmarks experience a drop in performance when using very aggressive thresholds (lower $\epsilon$). Despite this, for moderate values of $\epsilon$, the benchmark scores remain close to those of the main model, while achieving speedups between 10\% and 20\% depending on the model size.

Interestingly, we observe that smaller models in the Pythia suite, such as the 70M and 160M parameter variants, can even show improvements on certain benchmarks when early exits are used aggressively. This suggests that early exit mechanisms may help regularize smaller models or mitigate overfitting in some cases.

Beyond benchmark results, we also observe that the speedup achieved by early exit increases with the size of the model. Larger models benefit more from early exit, with greater computational savings for the same threshold values. This effect may be explained by several factors: (1) the relative size of the early exit heads becomes less significant as model size increases, and (2) the intermediate features in larger models may provide better representations, enabling more accurate early predictions. This trend is illustrated in Figure~\ref{fig:all_models_speedup_vs_threshold}, which shows the speedup as a function of the threshold for all model sizes.

All detailed tables of benchmark scores and speedup for each model and threshold are provided in the appendix.
While our results demonstrate the effectiveness of early exits, it is important to consider the limitations and broader impacts of this approach.

To address the accuracy drops observed with aggressive early exiting while maintaining computational efficiency, we adapt our calibrated early exit mechanism to speculative decoding. This approach allows us to achieve significant speedups without compromising the final output quality, as the full model verifies all predictions.

\subsection{Dynamic Early Exit for Speculative Decoding}
\label{subsec:dynamic_speculative_decoding}

Having established the effectiveness of calibrated early exits for single-token prediction, we now extend our method to \emph{speculative decoding}—a technique where a faster draft model proposes multiple tokens that are then verified in parallel by the full model. This approach, exemplified by LayerSkip~\cite{elhoushi2024layerskip}, can significantly accelerate autoregressive generation. However, LayerSkip requires practitioners to manually tune two coupled hyperparameters: (i) which intermediate head/layer to use for drafting, and (ii) how many tokens to speculate per round. Finding optimal settings often requires exhaustive grid search across prompts and tasks, with 24 configurations tested in our experiments.

Our key contribution is to replace these two discrete knobs with a \emph{single continuous accuracy threshold} $\epsilon\in[0,1]$, leveraging the same calibrated entropy-based exit mechanism from Section~\ref{subsubsec:calibration}. The key insight is that per-token confidence naturally determines both \emph{where} to exit (head selection) and \emph{when} to stop drafting (adaptive speculation length). This unified approach, which we call \textbf{Dynamic Self-Speculative Decoding (DSSD)}, eliminates manual tuning while achieving superior acceptance rates: on Pythia-2.8B, DSSD reaches \textbf{88.8\% acceptance at $\epsilon=0.9$} compared to LayerSkip's best of 53.6\%, a \textbf{1.66$\times$ improvement}.

\subsubsection{DSSD Algorithm}

DSSD extends the inference procedure from Section~\ref{subsubsec:inferenceprocess} to multi-token drafting:

\paragraph{Drafting phase.}
Starting from the current context, we evaluate early-exit heads sequentially from shallowest to deepest. For each position, we select the first head $k$ whose entropy $H(\mathbf{p}_k) < \tau_k(\epsilon)$ falls below its calibrated threshold. We append the predicted token to a draft buffer and continue until: (i) the buffer reaches a preset limit (e.g., 32 tokens), (ii) an end-of-sequence token is generated, or (iii) no head meets the confidence threshold, triggering a fallback to the full model. This produces a variable-length draft sequence whose length adapts to local token difficulty.

\paragraph{Verification phase.}
We pass the entire draft buffer through the full model in a single forward pass, obtaining target predictions for all positions simultaneously. We then compare each drafted token with the corresponding full-model prediction: accepted tokens are committed to the output, while the first mismatch triggers an immediate rewrite with the full-model token. This produces an \emph{atomic commit} of the accepted prefix. The mean number of accepted tokens per verification round acts as an implicit, adaptive speculation budget.

\paragraph{Comparison to LayerSkip.}
In LayerSkip, practitioners fix a single head (e.g., layer 6) and a speculation length (e.g., 10 tokens) for all contexts. Our method dynamically chooses the head on a per-token basis and automatically adjusts the effective speculation length through the acceptance mechanism, adapting to context difficulty without manual intervention.

\subsubsection{Experimental Design}

We evaluate DSSD on Pythia-2.8B with 200-token greedy decoding, comparing:

\begin{itemize}
    \item \textbf{DSSD (ours)}: Accuracy sweep $\epsilon\in\{0.5, 0.55, 0.6, 0.65, 0.7, 0.8, 0.9\}$ (7 levels).
    \item \textbf{LayerSkip (fixed)}: Exhaustive grid search over 4 heads $\times$ 6 speculation lengths (24 configurations).
\end{itemize}

Experiments use prompts sampled from our dataset to test the method across diverse topics.

\begin{figure}[t]
    \centering
    \includegraphics[width=\columnwidth]{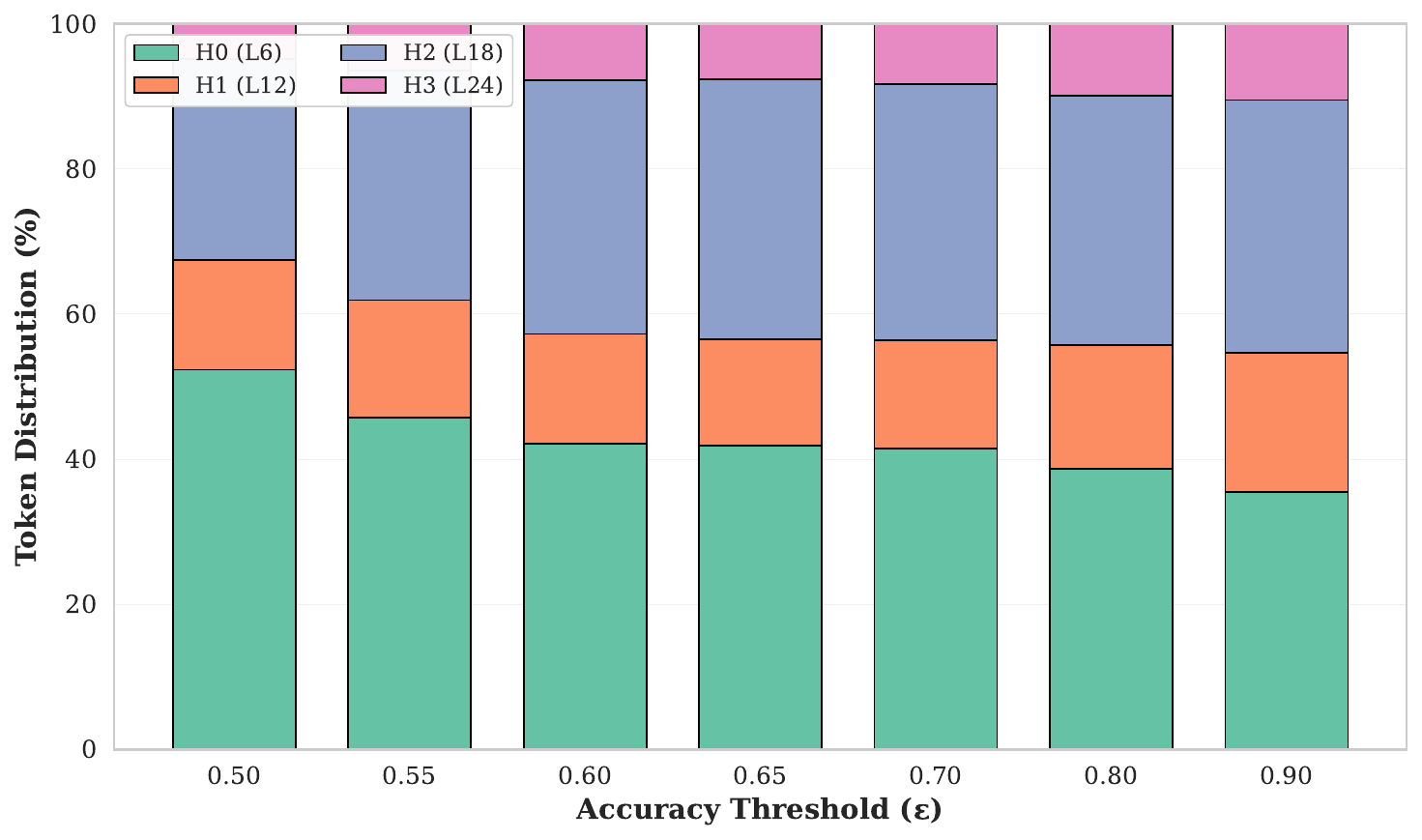}
    \caption{Adaptive head selection: token distribution across heads H0(L6), H1(L12), H2(L18), H3(L24) at different accuracy thresholds $\epsilon$, where notation $H_i(L_j)$ indicates head $i$ at layer $j$. As $\epsilon$ decreases, the system shifts from primarily using H0 to a balanced mix across all heads.}
    \label{fig:head_usage_dynamic}
\end{figure}

\subsubsection{Results and Analysis}

\paragraph{Superior acceptance with lower overhead.}
DSSD at $\epsilon=0.9$ achieves \textbf{88.8\% acceptance rate}—1.66$\times$ better than LayerSkip's best (53.6\%). This improvement translates directly to reduced wasted computation: DSSD discards only 8 tokens versus LayerSkip's 128 tokens, a \textbf{14$\times$ reduction} in waste. Figure~\ref{fig:head_usage_dynamic} illustrates this adaptive behavior.

%\begin{figure}[t]
%    \centering
%    \includegraphics[width=\columnwidth]{paper_figure2_comparison.pdf}
%    \caption{Comprehensive comparison of 31 configurations: 7 DSSD (diamonds with $\epsilon$ labels) vs 24 LayerSkip (circles). Panel (a) plots acceptance rate against mean exit layer (computational cost). DSSD consistently achieves higher acceptance while using moderate depth. Gold/silver stars mark best configurations. Panel (b) shows wasted tokens vs mean layer. DSSD wastes 14$\times$ fewer tokens (9 vs 128).}
%    \label{fig:dynamic_vs_layerskip_comparison}
%\end{figure}

\begin{figure}[t]
    \centering
    \includegraphics[width=0.95\columnwidth]{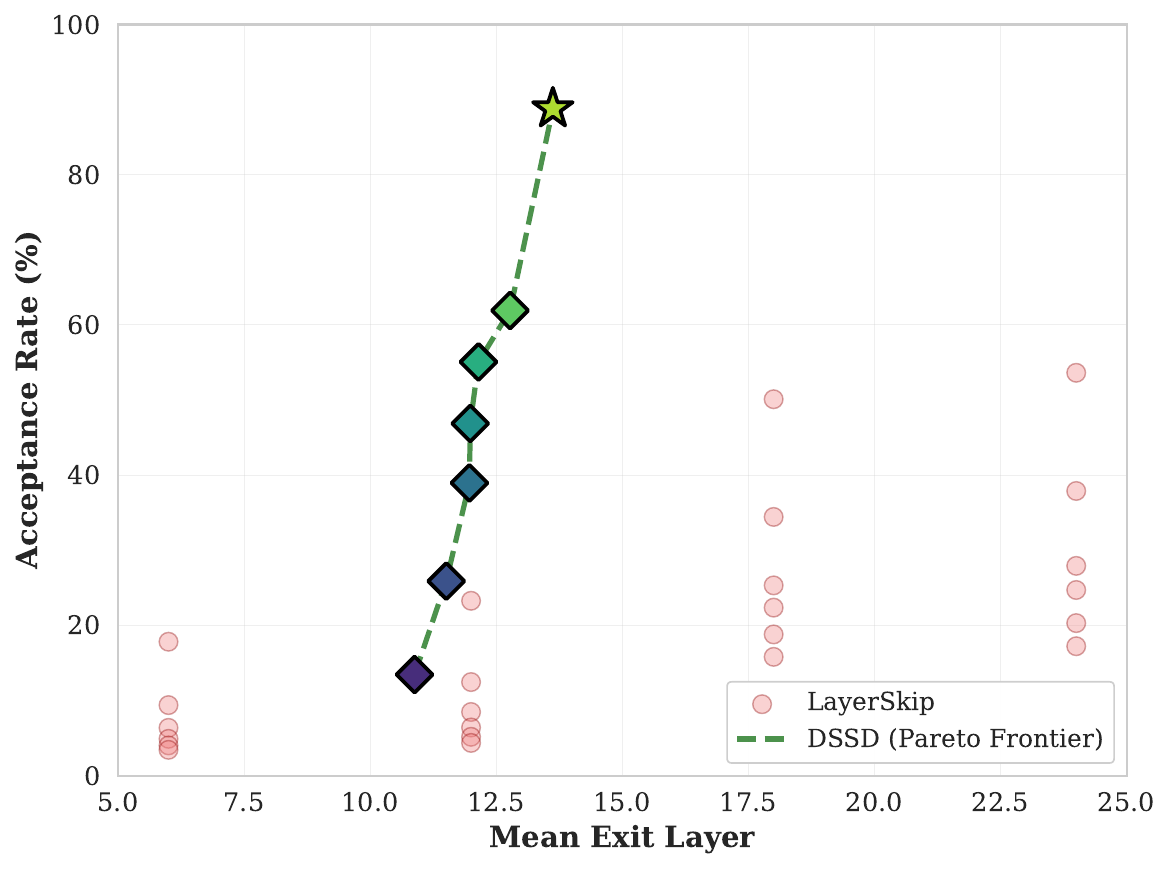}
    \caption{Acceptance rate vs mean exit layer during drafting across all configurations. The dashed green line traces DSSD's frontier (diamonds), while LayerSkip configurations (circles) are dominated. The star highlights DSSD at $\epsilon=0.9$, achieving 88\% acceptance with only 13.6 mean exit layers.}
    \label{fig:pareto_frontier}
\end{figure}

\paragraph{Understanding layer concentration.}
An important observation from Figure~\ref{fig:head_usage_dynamic}b is that mean exit layers concentrate in a relatively narrow range (10.9--13.6 across all $\epsilon$ values), spanning only 2.74 layers despite having 4 discrete exit points at [6, 12, 18, 24]. This concentration is \textbf{expected and optimal} rather than anomalous: (i) only 4 discrete exit points naturally limit the range, (ii) calibrated thresholds steer tokens toward heads H1-H2 (layers 12-18) as these offer the best efficiency-quality trade-off, and (iii) the narrow span represents successful optimization—the system has identified that most tokens benefit from moderate depth.

\subsubsection{Summary}

Our dynamic early-exit controller unifies head selection and speculation length into a single accuracy threshold $\epsilon$, eliminating exhaustive hyperparameter tuning. Across 31 configurations tested on Pythia-2.8B, our method demonstrates:

\begin{itemize}
    \item \textbf{Superior acceptance}: 1.66$\times$ higher than best LayerSkip (88.8\% vs 53.6\%)
    \item \textbf{Reduced waste}: 14$\times$ fewer discarded tokens (9 vs 128)
    \item \textbf{Earlier exit}: 10.4 layers shallower on average (13.6 vs 24.0)
    \item \textbf{Zero tuning}: Single threshold vs 2D grid search
\end{itemize}

As model size increases and exhaustive search becomes prohibitive, the automatic adaptation to token difficulty positions DSSD as a scalable, user-friendly alternative to fixed speculative strategies.

\section{Limitations and Potential Impacts}

While our early exit approach enables acceleration of large language models, the speedup gains without DSSD can remain modest when aiming to preserve the original LLM's accuracy. However, we observe that for certain benchmarks, early exits do not lead to any noticeable degradation in accuracy, even with significant acceleration.

Another important trend highlighted by our experiments is that the speedup obtained through early exits tends to increase with model size. Larger models appear to benefit more from this mechanism, both in terms of computational savings and in the stability of their predictions under early exit. Nevertheless, due to our limited computational resources, we were unable to train and evaluate early exit models on the largest architectures available. Confirming and further exploring this trend would require access to greater computing power.

\section{Conclusion}
\label{sec:conclusion}

In this work, we introduced a modular early exit mechanism for large language models, allowing inference to terminate early based on calibrated confidence metrics. Our approach is easy to integrate into existing transformer architectures and does not require retraining the backbone model. Through extensive experiments on the Pythia suite, we demonstrated that early exits can provide significant inference speedups, especially for larger models, while maintaining high accuracy on several benchmarks.

Furthermore, we introduced Dynamic Self-Speculative Decoding (DSSD), which integrates calibrated heads into a speculative decoder to achieve 1.66$\times$ higher token acceptance rates compared to manually-tuned LayerSkip baselines. The zero-tuning property and automatic adaptation to token difficulty make DSSD particularly attractive for practical deployment.

Overall, our findings suggest that exploiting the natural variability in token difficulty is a promising direction for accelerating large language models. We encourage the community to build upon this work, exploring new strategies and applications to enable efficient and scalable deployment of LLMs in real-world, resource-constrained environments.

\bibliographystyle{plainnat}
\bibliography{biblio}

\onecolumn
\appendix
\appendix

\section{Appendix}
\label{sec:appendix}
\subsection{Full benchmark results}
\label{subsec:fullbenchmarkresults}

This appendix presents the complete benchmark results for all Pythia models evaluated in this work. The following tables report the benchmark scores for each model size (from 70M to 2.8B parameters) on a range of tasks (WSC, Winogrande, SciQ, PIQA, LogiQA, LAMBADA\_OpenAI, ARC Easy, ARC Challenge) for different values of the confidence threshold $\epsilon$. For each configuration, we also indicate the corresponding speedup achieved by the early exit mechanism. These results allow a detailed comparison of the trade-off between computational savings and accuracy across models and tasks.

\begin{table}[htbp!]
\footnotesize
\centering
\caption{Results for Pythia-70M (Part 1/2)}
\begin{tabular}{lcccccc}
\hline
$\epsilon$ & Speedup & wsc & winogrande & sciq & piqa & logiqa \\
\hline
0.50 & {\color[RGB]{0,149,0} 24.0\%} & {\color[RGB]{0,0,0} 0.635} $\pm$ {\color[RGB]{0,0,0} 0.047} & {\color[RGB]{0,0,0} 0.522} $\pm$ {\color[RGB]{0,0,0} 0.014} & {\color[RGB]{0,255,0} 0.320} $\pm$ {\color[RGB]{0,0,0} 0.015} & {\color[RGB]{0,0,0} 0.557} $\pm$ {\color[RGB]{0,0,0} 0.012} & {\color[RGB]{0,0,0} 0.206} $\pm$ {\color[RGB]{0,0,0} 0.016} \\
0.55 & {\color[RGB]{0,139,0} 19.2\%} & {\color[RGB]{255,0,0} 0.442} $\pm$ {\color[RGB]{0,0,0} 0.049} & {\color[RGB]{0,0,0} 0.522} $\pm$ {\color[RGB]{0,0,0} 0.014} & {\color[RGB]{0,255,0} 0.310} $\pm$ {\color[RGB]{0,0,0} 0.015} & {\color[RGB]{0,0,0} 0.550} $\pm$ {\color[RGB]{0,0,0} 0.012} & {\color[RGB]{0,0,0} 0.204} $\pm$ {\color[RGB]{0,0,0} 0.016} \\
0.60 & {\color[RGB]{0,131,0} 15.4\%} & {\color[RGB]{0,0,0} 0.615} $\pm$ {\color[RGB]{0,0,0} 0.048} & {\color[RGB]{0,0,0} 0.520} $\pm$ {\color[RGB]{0,0,0} 0.014} & {\color[RGB]{0,255,0} 0.310} $\pm$ {\color[RGB]{0,0,0} 0.015} & {\color[RGB]{0,0,0} 0.557} $\pm$ {\color[RGB]{0,0,0} 0.012} & {\color[RGB]{0,0,0} 0.210} $\pm$ {\color[RGB]{0,0,0} 0.016} \\
0.65 & {\color[RGB]{0,124,0} 12.0\%} & {\color[RGB]{0,0,0} 0.606} $\pm$ {\color[RGB]{0,0,0} 0.048} & {\color[RGB]{0,0,0} 0.525} $\pm$ {\color[RGB]{0,0,0} 0.014} & {\color[RGB]{0,248,0} 0.279} $\pm$ {\color[RGB]{0,0,0} 0.014} & {\color[RGB]{0,0,0} 0.552} $\pm$ {\color[RGB]{0,0,0} 0.012} & {\color[RGB]{0,0,0} 0.209} $\pm$ {\color[RGB]{0,0,0} 0.016} \\
0.70 & {\color[RGB]{0,119,0} 9.4\%} & {\color[RGB]{0,0,0} 0.606} $\pm$ {\color[RGB]{0,0,0} 0.048} & {\color[RGB]{0,0,0} 0.530} $\pm$ {\color[RGB]{0,0,0} 0.014} & {\color[RGB]{0,202,0} 0.264} $\pm$ {\color[RGB]{0,0,0} 0.014} & {\color[RGB]{0,0,0} 0.552} $\pm$ {\color[RGB]{0,0,0} 0.012} & {\color[RGB]{0,0,0} 0.198} $\pm$ {\color[RGB]{0,0,0} 0.016} \\
0.75 & {\color[RGB]{0,115,0} 7.4\%} & {\color[RGB]{0,0,0} 0.606} $\pm$ {\color[RGB]{0,0,0} 0.048} & {\color[RGB]{0,0,0} 0.521} $\pm$ {\color[RGB]{0,0,0} 0.014} & {\color[RGB]{0,0,0} 0.242} $\pm$ {\color[RGB]{0,0,0} 0.014} & {\color[RGB]{0,0,0} 0.548} $\pm$ {\color[RGB]{0,0,0} 0.012} & {\color[RGB]{0,0,0} 0.201} $\pm$ {\color[RGB]{0,0,0} 0.016} \\
0.80 & {\color[RGB]{0,110,0} 5.2\%} & {\color[RGB]{0,0,0} 0.606} $\pm$ {\color[RGB]{0,0,0} 0.048} & {\color[RGB]{0,0,0} 0.528} $\pm$ {\color[RGB]{0,0,0} 0.014} & {\color[RGB]{0,0,0} 0.231} $\pm$ {\color[RGB]{0,0,0} 0.013} & {\color[RGB]{0,0,0} 0.548} $\pm$ {\color[RGB]{0,0,0} 0.012} & {\color[RGB]{0,0,0} 0.195} $\pm$ {\color[RGB]{0,0,0} 0.016} \\
0.85 & {\color[RGB]{0,107,0} 3.7\%} & {\color[RGB]{0,0,0} 0.606} $\pm$ {\color[RGB]{0,0,0} 0.048} & {\color[RGB]{0,0,0} 0.529} $\pm$ {\color[RGB]{0,0,0} 0.014} & {\color[RGB]{0,0,0} 0.231} $\pm$ {\color[RGB]{0,0,0} 0.013} & {\color[RGB]{0,0,0} 0.548} $\pm$ {\color[RGB]{0,0,0} 0.012} & {\color[RGB]{0,0,0} 0.198} $\pm$ {\color[RGB]{0,0,0} 0.016} \\
0.90 & {\color[RGB]{0,104,0} 2.1\%} & {\color[RGB]{0,0,0} 0.606} $\pm$ {\color[RGB]{0,0,0} 0.048} & {\color[RGB]{0,0,0} 0.524} $\pm$ {\color[RGB]{0,0,0} 0.014} & {\color[RGB]{0,0,0} 0.230} $\pm$ {\color[RGB]{0,0,0} 0.013} & {\color[RGB]{0,0,0} 0.548} $\pm$ {\color[RGB]{0,0,0} 0.012} & {\color[RGB]{0,0,0} 0.203} $\pm$ {\color[RGB]{0,0,0} 0.016} \\
0.95 & {\color[RGB]{0,101,0} 0.7\%} & {\color[RGB]{0,0,0} 0.606} $\pm$ {\color[RGB]{0,0,0} 0.048} & {\color[RGB]{0,0,0} 0.525} $\pm$ {\color[RGB]{0,0,0} 0.014} & {\color[RGB]{0,0,0} 0.231} $\pm$ {\color[RGB]{0,0,0} 0.013} & {\color[RGB]{0,0,0} 0.550} $\pm$ {\color[RGB]{0,0,0} 0.012} & {\color[RGB]{0,0,0} 0.212} $\pm$ {\color[RGB]{0,0,0} 0.016} \\
1.00 & {\color[RGB]{0,100,0} 0.0\%} & {\color[RGB]{0,0,0} 0.606} $\pm$ {\color[RGB]{0,0,0} 0.048} & {\color[RGB]{0,0,0} 0.526} $\pm$ {\color[RGB]{0,0,0} 0.014} & {\color[RGB]{0,0,0} 0.231} $\pm$ {\color[RGB]{0,0,0} 0.013} & {\color[RGB]{0,0,0} 0.550} $\pm$ {\color[RGB]{0,0,0} 0.012} & {\color[RGB]{0,0,0} 0.210} $\pm$ {\color[RGB]{0,0,0} 0.016} \\
\hline
\end{tabular}
\label{tab:pythia70m_part1}
\end{table}
\begin{table}[htbp!]
\centering
\caption{Results for Pythia-70M (Part 2/2)}
\begin{tabular}{lcccc}
\hline
$\epsilon$ & Speedup & lambada\_openai & arc\_easy & arc\_challenge \\
\hline
0.50 & {\color[RGB]{0,149,0} 24.0\%} & {\color[RGB]{0,0,0} 0.015} $\pm$ {\color[RGB]{0,0,0} 0.002} & {\color[RGB]{0,0,0} 0.291} $\pm$ {\color[RGB]{0,0,0} 0.009} & {\color[RGB]{0,139,0} 0.203} $\pm$ {\color[RGB]{0,0,0} 0.012} \\
0.55 & {\color[RGB]{0,139,0} 19.2\%} & {\color[RGB]{0,0,0} 0.016} $\pm$ {\color[RGB]{0,0,0} 0.002} & {\color[RGB]{0,0,0} 0.294} $\pm$ {\color[RGB]{0,0,0} 0.009} & {\color[RGB]{0,0,0} 0.197} $\pm$ {\color[RGB]{0,0,0} 0.012} \\
0.60 & {\color[RGB]{0,131,0} 15.4\%} & {\color[RGB]{0,0,0} 0.016} $\pm$ {\color[RGB]{0,0,0} 0.002} & {\color[RGB]{0,137,0} 0.308} $\pm$ {\color[RGB]{0,0,0} 0.009} & {\color[RGB]{0,0,0} 0.194} $\pm$ {\color[RGB]{0,0,0} 0.012} \\
0.65 & {\color[RGB]{0,124,0} 12.0\%} & {\color[RGB]{0,0,0} 0.016} $\pm$ {\color[RGB]{0,0,0} 0.002} & {\color[RGB]{0,0,0} 0.303} $\pm$ {\color[RGB]{0,0,0} 0.009} & {\color[RGB]{0,0,0} 0.193} $\pm$ {\color[RGB]{0,0,0} 0.012} \\
0.70 & {\color[RGB]{0,119,0} 9.4\%} & {\color[RGB]{0,0,0} 0.016} $\pm$ {\color[RGB]{0,0,0} 0.002} & {\color[RGB]{0,0,0} 0.302} $\pm$ {\color[RGB]{0,0,0} 0.009} & {\color[RGB]{0,0,0} 0.183} $\pm$ {\color[RGB]{0,0,0} 0.011} \\
0.75 & {\color[RGB]{0,115,0} 7.4\%} & {\color[RGB]{0,0,0} 0.016} $\pm$ {\color[RGB]{0,0,0} 0.002} & {\color[RGB]{0,0,0} 0.298} $\pm$ {\color[RGB]{0,0,0} 0.009} & {\color[RGB]{0,0,0} 0.191} $\pm$ {\color[RGB]{0,0,0} 0.011} \\
0.80 & {\color[RGB]{0,110,0} 5.2\%} & {\color[RGB]{0,0,0} 0.016} $\pm$ {\color[RGB]{0,0,0} 0.002} & {\color[RGB]{0,0,0} 0.295} $\pm$ {\color[RGB]{0,0,0} 0.009} & {\color[RGB]{0,0,0} 0.193} $\pm$ {\color[RGB]{0,0,0} 0.012} \\
0.85 & {\color[RGB]{0,107,0} 3.7\%} & {\color[RGB]{0,0,0} 0.016} $\pm$ {\color[RGB]{0,0,0} 0.002} & {\color[RGB]{0,0,0} 0.295} $\pm$ {\color[RGB]{0,0,0} 0.009} & {\color[RGB]{0,0,0} 0.193} $\pm$ {\color[RGB]{0,0,0} 0.012} \\
0.90 & {\color[RGB]{0,104,0} 2.1\%} & {\color[RGB]{0,0,0} 0.016} $\pm$ {\color[RGB]{0,0,0} 0.002} & {\color[RGB]{0,0,0} 0.296} $\pm$ {\color[RGB]{0,0,0} 0.009} & {\color[RGB]{0,0,0} 0.191} $\pm$ {\color[RGB]{0,0,0} 0.011} \\
0.95 & {\color[RGB]{0,101,0} 0.7\%} & {\color[RGB]{0,0,0} 0.016} $\pm$ {\color[RGB]{0,0,0} 0.002} & {\color[RGB]{0,0,0} 0.296} $\pm$ {\color[RGB]{0,0,0} 0.009} & {\color[RGB]{0,0,0} 0.190} $\pm$ {\color[RGB]{0,0,0} 0.011} \\
1.00 & {\color[RGB]{0,100,0} 0.0\%} & {\color[RGB]{0,0,0} 0.016} $\pm$ {\color[RGB]{0,0,0} 0.002} & {\color[RGB]{0,0,0} 0.296} $\pm$ {\color[RGB]{0,0,0} 0.009} & {\color[RGB]{0,0,0} 0.190} $\pm$ {\color[RGB]{0,0,0} 0.011} \\
\hline
\end{tabular}
\label{tab:pythia70m_part2}
\end{table}
\begin{table}[htbp!]
\footnotesize
\centering
\caption{Results for Pythia-160M (Part 1/2)}
\begin{tabular}{lcccccc}
\hline
$\epsilon$ & Speedup & wsc & winogrande & sciq & piqa & logiqa \\
\hline
0.50 & {\color[RGB]{0,184,0} 40.7\%} & {\color[RGB]{255,0,0} 0.365} $\pm$ {\color[RGB]{0,0,0} 0.047} & {\color[RGB]{0,210,0} 0.519} $\pm$ {\color[RGB]{0,0,0} 0.014} & {\color[RGB]{255,0,0} 0.394} $\pm$ {\color[RGB]{0,0,0} 0.015} & {\color[RGB]{137,0,0} 0.569} $\pm$ {\color[RGB]{0,0,0} 0.012} & {\color[RGB]{0,0,0} 0.237} $\pm$ {\color[RGB]{0,0,0} 0.017} \\
0.55 & {\color[RGB]{0,173,0} 35.4\%} & {\color[RGB]{255,0,0} 0.365} $\pm$ {\color[RGB]{0,0,0} 0.047} & {\color[RGB]{0,192,0} 0.513} $\pm$ {\color[RGB]{0,0,0} 0.014} & {\color[RGB]{255,0,0} 0.427} $\pm$ {\color[RGB]{0,0,0} 0.016} & {\color[RGB]{0,0,0} 0.581} $\pm$ {\color[RGB]{0,0,0} 0.012} & {\color[RGB]{176,0,0} 0.204} $\pm$ {\color[RGB]{0,0,0} 0.016} \\
0.60 & {\color[RGB]{0,152,0} 25.6\%} & {\color[RGB]{255,0,0} 0.423} $\pm$ {\color[RGB]{0,0,0} 0.049} & {\color[RGB]{0,197,0} 0.515} $\pm$ {\color[RGB]{0,0,0} 0.014} & {\color[RGB]{220,0,0} 0.459} $\pm$ {\color[RGB]{0,0,0} 0.016} & {\color[RGB]{0,0,0} 0.578} $\pm$ {\color[RGB]{0,0,0} 0.012} & {\color[RGB]{157,0,0} 0.210} $\pm$ {\color[RGB]{0,0,0} 0.016} \\
0.65 & {\color[RGB]{0,143,0} 21.0\%} & {\color[RGB]{255,0,0} 0.500} $\pm$ {\color[RGB]{0,0,0} 0.049} & {\color[RGB]{0,156,0} 0.501} $\pm$ {\color[RGB]{0,0,0} 0.014} & {\color[RGB]{174,0,0} 0.474} $\pm$ {\color[RGB]{0,0,0} 0.016} & {\color[RGB]{0,0,0} 0.587} $\pm$ {\color[RGB]{0,0,0} 0.011} & {\color[RGB]{0,0,0} 0.217} $\pm$ {\color[RGB]{0,0,0} 0.016} \\
0.70 & {\color[RGB]{0,135,0} 17.1\%} & {\color[RGB]{0,0,0} 0.567} $\pm$ {\color[RGB]{0,0,0} 0.049} & {\color[RGB]{0,148,0} 0.499} $\pm$ {\color[RGB]{0,0,0} 0.014} & {\color[RGB]{162,0,0} 0.478} $\pm$ {\color[RGB]{0,0,0} 0.016} & {\color[RGB]{0,0,0} 0.579} $\pm$ {\color[RGB]{0,0,0} 0.012} & {\color[RGB]{176,0,0} 0.204} $\pm$ {\color[RGB]{0,0,0} 0.016} \\
0.75 & {\color[RGB]{0,127,0} 13.2\%} & {\color[RGB]{0,0,0} 0.567} $\pm$ {\color[RGB]{0,0,0} 0.049} & {\color[RGB]{0,0,0} 0.493} $\pm$ {\color[RGB]{0,0,0} 0.014} & {\color[RGB]{158,0,0} 0.479} $\pm$ {\color[RGB]{0,0,0} 0.016} & {\color[RGB]{0,0,0} 0.585} $\pm$ {\color[RGB]{0,0,0} 0.011} & {\color[RGB]{180,0,0} 0.203} $\pm$ {\color[RGB]{0,0,0} 0.016} \\
0.80 & {\color[RGB]{0,121,0} 10.4\%} & {\color[RGB]{0,0,0} 0.567} $\pm$ {\color[RGB]{0,0,0} 0.049} & {\color[RGB]{0,0,0} 0.489} $\pm$ {\color[RGB]{0,0,0} 0.014} & {\color[RGB]{149,0,0} 0.482} $\pm$ {\color[RGB]{0,0,0} 0.016} & {\color[RGB]{0,0,0} 0.582} $\pm$ {\color[RGB]{0,0,0} 0.012} & {\color[RGB]{0,0,0} 0.218} $\pm$ {\color[RGB]{0,0,0} 0.016} \\
0.85 & {\color[RGB]{0,116,0} 8.1\%} & {\color[RGB]{0,0,0} 0.567} $\pm$ {\color[RGB]{0,0,0} 0.049} & {\color[RGB]{0,0,0} 0.485} $\pm$ {\color[RGB]{0,0,0} 0.014} & {\color[RGB]{0,0,0} 0.492} $\pm$ {\color[RGB]{0,0,0} 0.016} & {\color[RGB]{0,0,0} 0.578} $\pm$ {\color[RGB]{0,0,0} 0.012} & {\color[RGB]{0,0,0} 0.220} $\pm$ {\color[RGB]{0,0,0} 0.016} \\
0.90 & {\color[RGB]{0,111,0} 5.7\%} & {\color[RGB]{0,0,0} 0.567} $\pm$ {\color[RGB]{0,0,0} 0.049} & {\color[RGB]{0,0,0} 0.486} $\pm$ {\color[RGB]{0,0,0} 0.014} & {\color[RGB]{0,0,0} 0.495} $\pm$ {\color[RGB]{0,0,0} 0.016} & {\color[RGB]{0,0,0} 0.581} $\pm$ {\color[RGB]{0,0,0} 0.012} & {\color[RGB]{0,0,0} 0.233} $\pm$ {\color[RGB]{0,0,0} 0.017} \\
0.95 & {\color[RGB]{0,106,0} 3.0\%} & {\color[RGB]{0,0,0} 0.567} $\pm$ {\color[RGB]{0,0,0} 0.049} & {\color[RGB]{0,0,0} 0.483} $\pm$ {\color[RGB]{0,0,0} 0.014} & {\color[RGB]{0,0,0} 0.499} $\pm$ {\color[RGB]{0,0,0} 0.016} & {\color[RGB]{0,0,0} 0.582} $\pm$ {\color[RGB]{0,0,0} 0.012} & {\color[RGB]{0,0,0} 0.227} $\pm$ {\color[RGB]{0,0,0} 0.016} \\
1.00 & {\color[RGB]{0,0,0} 0.0\%} & {\color[RGB]{0,0,0} 0.567} $\pm$ {\color[RGB]{0,0,0} 0.049} & {\color[RGB]{0,0,0} 0.483} $\pm$ {\color[RGB]{0,0,0} 0.014} & {\color[RGB]{0,0,0} 0.498} $\pm$ {\color[RGB]{0,0,0} 0.016} & {\color[RGB]{0,0,0} 0.581} $\pm$ {\color[RGB]{0,0,0} 0.012} & {\color[RGB]{0,0,0} 0.229} $\pm$ {\color[RGB]{0,0,0} 0.016} \\
\hline
\end{tabular}
\label{tab:pythia160m_part1}
\end{table}
\begin{table}[htbp!]
\centering
\caption{Results for Pythia-160M (Part 2/2)}
\begin{tabular}{lcccc}
\hline
$\epsilon$ & Speedup & lambada\_openai & arc\_easy & arc\_challenge \\
\hline
0.50 & {\color[RGB]{0,184,0} 40.7\%} & {\color[RGB]{245,0,0} 0.081} $\pm$ {\color[RGB]{0,0,0} 0.004} & {\color[RGB]{255,0,0} 0.307} $\pm$ {\color[RGB]{0,0,0} 0.009} & {\color[RGB]{0,0,0} 0.209} $\pm$ {\color[RGB]{0,0,0} 0.012} \\
0.55 & {\color[RGB]{0,173,0} 35.4\%} & {\color[RGB]{216,0,0} 0.090} $\pm$ {\color[RGB]{0,0,0} 0.004} & {\color[RGB]{255,0,0} 0.321} $\pm$ {\color[RGB]{0,0,0} 0.010} & {\color[RGB]{0,0,0} 0.209} $\pm$ {\color[RGB]{0,0,0} 0.012} \\
0.60 & {\color[RGB]{0,152,0} 25.6\%} & {\color[RGB]{186,0,0} 0.100} $\pm$ {\color[RGB]{0,0,0} 0.004} & {\color[RGB]{206,0,0} 0.337} $\pm$ {\color[RGB]{0,0,0} 0.010} & {\color[RGB]{0,0,0} 0.201} $\pm$ {\color[RGB]{0,0,0} 0.012} \\
0.65 & {\color[RGB]{0,143,0} 21.0\%} & {\color[RGB]{163,0,0} 0.107} $\pm$ {\color[RGB]{0,0,0} 0.004} & {\color[RGB]{143,0,0} 0.357} $\pm$ {\color[RGB]{0,0,0} 0.010} & {\color[RGB]{0,0,0} 0.197} $\pm$ {\color[RGB]{0,0,0} 0.012} \\
0.70 & {\color[RGB]{0,135,0} 17.1\%} & {\color[RGB]{148,0,0} 0.112} $\pm$ {\color[RGB]{0,0,0} 0.004} & {\color[RGB]{0,0,0} 0.362} $\pm$ {\color[RGB]{0,0,0} 0.010} & {\color[RGB]{137,0,0} 0.194} $\pm$ {\color[RGB]{0,0,0} 0.012} \\
0.75 & {\color[RGB]{0,127,0} 13.2\%} & {\color[RGB]{135,0,0} 0.116} $\pm$ {\color[RGB]{0,0,0} 0.004} & {\color[RGB]{0,0,0} 0.366} $\pm$ {\color[RGB]{0,0,0} 0.010} & {\color[RGB]{147,0,0} 0.190} $\pm$ {\color[RGB]{0,0,0} 0.011} \\
0.80 & {\color[RGB]{0,121,0} 10.4\%} & {\color[RGB]{128,0,0} 0.118} $\pm$ {\color[RGB]{0,0,0} 0.004} & {\color[RGB]{0,0,0} 0.371} $\pm$ {\color[RGB]{0,0,0} 0.010} & {\color[RGB]{0,0,0} 0.195} $\pm$ {\color[RGB]{0,0,0} 0.012} \\
0.85 & {\color[RGB]{0,116,0} 8.1\%} & {\color[RGB]{122,0,0} 0.120} $\pm$ {\color[RGB]{0,0,0} 0.005} & {\color[RGB]{0,0,0} 0.371} $\pm$ {\color[RGB]{0,0,0} 0.010} & {\color[RGB]{0,0,0} 0.203} $\pm$ {\color[RGB]{0,0,0} 0.012} \\
0.90 & {\color[RGB]{0,111,0} 5.7\%} & {\color[RGB]{0,0,0} 0.124} $\pm$ {\color[RGB]{0,0,0} 0.005} & {\color[RGB]{0,0,0} 0.372} $\pm$ {\color[RGB]{0,0,0} 0.010} & {\color[RGB]{0,0,0} 0.204} $\pm$ {\color[RGB]{0,0,0} 0.012} \\
0.95 & {\color[RGB]{0,106,0} 3.0\%} & {\color[RGB]{0,0,0} 0.127} $\pm$ {\color[RGB]{0,0,0} 0.005} & {\color[RGB]{0,0,0} 0.371} $\pm$ {\color[RGB]{0,0,0} 0.010} & {\color[RGB]{0,0,0} 0.206} $\pm$ {\color[RGB]{0,0,0} 0.012} \\
1.00 & {\color[RGB]{0,0,0} 0.0\%} & {\color[RGB]{0,0,0} 0.127} $\pm$ {\color[RGB]{0,0,0} 0.005} & {\color[RGB]{0,0,0} 0.371} $\pm$ {\color[RGB]{0,0,0} 0.010} & {\color[RGB]{0,0,0} 0.206} $\pm$ {\color[RGB]{0,0,0} 0.012} \\
\hline
\end{tabular}
\label{tab:pythia160m_part2}
\end{table}
\begin{table}[htbp!]
\footnotesize
\centering
\caption{Results for Pythia-410M (Part 1/2)}
\begin{tabular}{lcccccc}
\hline
$\epsilon$ & Speedup & wsc & winogrande & sciq & piqa & logiqa \\
\hline
0.50 & {\color[RGB]{0,201,0} 49.2\%} & {\color[RGB]{255,0,0} 0.500} $\pm$ {\color[RGB]{0,0,0} 0.049} & {\color[RGB]{0,0,0} 0.522} $\pm$ {\color[RGB]{0,0,0} 0.014} & {\color[RGB]{255,0,0} 0.622} $\pm$ {\color[RGB]{0,0,0} 0.015} & {\color[RGB]{255,0,0} 0.606} $\pm$ {\color[RGB]{0,0,0} 0.011} & {\color[RGB]{0,0,0} 0.217} $\pm$ {\color[RGB]{0,0,0} 0.016} \\
0.55 & {\color[RGB]{0,184,0} 40.8\%} & {\color[RGB]{0,0,0} 0.567} $\pm$ {\color[RGB]{0,0,0} 0.049} & {\color[RGB]{158,0,0} 0.511} $\pm$ {\color[RGB]{0,0,0} 0.014} & {\color[RGB]{255,0,0} 0.699} $\pm$ {\color[RGB]{0,0,0} 0.015} & {\color[RGB]{236,0,0} 0.628} $\pm$ {\color[RGB]{0,0,0} 0.011} & {\color[RGB]{176,0,0} 0.201} $\pm$ {\color[RGB]{0,0,0} 0.016} \\
0.60 & {\color[RGB]{0,170,0} 34.2\%} & {\color[RGB]{0,0,0} 0.596} $\pm$ {\color[RGB]{0,0,0} 0.048} & {\color[RGB]{144,0,0} 0.516} $\pm$ {\color[RGB]{0,0,0} 0.014} & {\color[RGB]{255,0,0} 0.756} $\pm$ {\color[RGB]{0,0,0} 0.014} & {\color[RGB]{204,0,0} 0.639} $\pm$ {\color[RGB]{0,0,0} 0.011} & {\color[RGB]{0,0,0} 0.220} $\pm$ {\color[RGB]{0,0,0} 0.016} \\
0.65 & {\color[RGB]{0,153,0} 25.9\%} & {\color[RGB]{0,0,0} 0.596} $\pm$ {\color[RGB]{0,0,0} 0.048} & {\color[RGB]{0,0,0} 0.527} $\pm$ {\color[RGB]{0,0,0} 0.014} & {\color[RGB]{217,0,0} 0.774} $\pm$ {\color[RGB]{0,0,0} 0.013} & {\color[RGB]{202,0,0} 0.639} $\pm$ {\color[RGB]{0,0,0} 0.011} & {\color[RGB]{0,0,0} 0.218} $\pm$ {\color[RGB]{0,0,0} 0.016} \\
0.70 & {\color[RGB]{0,144,0} 21.3\%} & {\color[RGB]{0,0,0} 0.596} $\pm$ {\color[RGB]{0,0,0} 0.048} & {\color[RGB]{0,0,0} 0.517} $\pm$ {\color[RGB]{0,0,0} 0.014} & {\color[RGB]{189,0,0} 0.783} $\pm$ {\color[RGB]{0,0,0} 0.013} & {\color[RGB]{155,0,0} 0.655} $\pm$ {\color[RGB]{0,0,0} 0.011} & {\color[RGB]{0,0,0} 0.212} $\pm$ {\color[RGB]{0,0,0} 0.016} \\
0.75 & {\color[RGB]{0,135,0} 17.1\%} & {\color[RGB]{0,0,0} 0.596} $\pm$ {\color[RGB]{0,0,0} 0.048} & {\color[RGB]{0,0,0} 0.528} $\pm$ {\color[RGB]{0,0,0} 0.014} & {\color[RGB]{162,0,0} 0.792} $\pm$ {\color[RGB]{0,0,0} 0.013} & {\color[RGB]{143,0,0} 0.658} $\pm$ {\color[RGB]{0,0,0} 0.011} & {\color[RGB]{0,0,0} 0.221} $\pm$ {\color[RGB]{0,0,0} 0.016} \\
0.80 & {\color[RGB]{0,128,0} 13.6\%} & {\color[RGB]{0,0,0} 0.596} $\pm$ {\color[RGB]{0,0,0} 0.048} & {\color[RGB]{0,0,0} 0.534} $\pm$ {\color[RGB]{0,0,0} 0.014} & {\color[RGB]{162,0,0} 0.792} $\pm$ {\color[RGB]{0,0,0} 0.013} & {\color[RGB]{138,0,0} 0.660} $\pm$ {\color[RGB]{0,0,0} 0.011} & {\color[RGB]{0,0,0} 0.229} $\pm$ {\color[RGB]{0,0,0} 0.016} \\
0.85 & {\color[RGB]{0,121,0} 10.2\%} & {\color[RGB]{0,0,0} 0.596} $\pm$ {\color[RGB]{0,0,0} 0.048} & {\color[RGB]{0,0,0} 0.526} $\pm$ {\color[RGB]{0,0,0} 0.014} & {\color[RGB]{0,0,0} 0.800} $\pm$ {\color[RGB]{0,0,0} 0.013} & {\color[RGB]{0,0,0} 0.669} $\pm$ {\color[RGB]{0,0,0} 0.011} & {\color[RGB]{0,0,0} 0.226} $\pm$ {\color[RGB]{0,0,0} 0.016} \\
0.90 & {\color[RGB]{0,115,0} 7.5\%} & {\color[RGB]{0,0,0} 0.596} $\pm$ {\color[RGB]{0,0,0} 0.048} & {\color[RGB]{0,0,0} 0.529} $\pm$ {\color[RGB]{0,0,0} 0.014} & {\color[RGB]{0,0,0} 0.803} $\pm$ {\color[RGB]{0,0,0} 0.013} & {\color[RGB]{0,0,0} 0.669} $\pm$ {\color[RGB]{0,0,0} 0.011} & {\color[RGB]{0,0,0} 0.214} $\pm$ {\color[RGB]{0,0,0} 0.016} \\
0.95 & {\color[RGB]{0,108,0} 4.3\%} & {\color[RGB]{0,0,0} 0.596} $\pm$ {\color[RGB]{0,0,0} 0.048} & {\color[RGB]{0,0,0} 0.532} $\pm$ {\color[RGB]{0,0,0} 0.014} & {\color[RGB]{0,0,0} 0.809} $\pm$ {\color[RGB]{0,0,0} 0.012} & {\color[RGB]{0,0,0} 0.671} $\pm$ {\color[RGB]{0,0,0} 0.011} & {\color[RGB]{0,0,0} 0.220} $\pm$ {\color[RGB]{0,0,0} 0.016} \\
1.00 & {\color[RGB]{0,0,0} 0.0\%} & {\color[RGB]{0,0,0} 0.596} $\pm$ {\color[RGB]{0,0,0} 0.048} & {\color[RGB]{0,0,0} 0.530} $\pm$ {\color[RGB]{0,0,0} 0.014} & {\color[RGB]{0,0,0} 0.812} $\pm$ {\color[RGB]{0,0,0} 0.012} & {\color[RGB]{0,0,0} 0.672} $\pm$ {\color[RGB]{0,0,0} 0.011} & {\color[RGB]{0,0,0} 0.226} $\pm$ {\color[RGB]{0,0,0} 0.016} \\
\hline
\end{tabular}
\label{tab:pythia410m_part1}
\end{table}
\begin{table}[htbp!]
\centering
\caption{Results for Pythia-410M (Part 2/2)}
\begin{tabular}{lcccc}
\hline
$\epsilon$ & Speedup & lambada\_openai & arc\_easy & arc\_challenge \\
\hline
0.50 & {\color[RGB]{0,201,0} 49.2\%} & {\color[RGB]{255,0,0} 0.234} $\pm$ {\color[RGB]{0,0,0} 0.006} & {\color[RGB]{255,0,0} 0.349} $\pm$ {\color[RGB]{0,0,0} 0.010} & {\color[RGB]{0,0,0} 0.212} $\pm$ {\color[RGB]{0,0,0} 0.012} \\
0.55 & {\color[RGB]{0,184,0} 40.8\%} & {\color[RGB]{255,0,0} 0.285} $\pm$ {\color[RGB]{0,0,0} 0.006} & {\color[RGB]{255,0,0} 0.388} $\pm$ {\color[RGB]{0,0,0} 0.010} & {\color[RGB]{0,0,0} 0.202} $\pm$ {\color[RGB]{0,0,0} 0.012} \\
0.60 & {\color[RGB]{0,170,0} 34.2\%} & {\color[RGB]{255,0,0} 0.332} $\pm$ {\color[RGB]{0,0,0} 0.007} & {\color[RGB]{255,0,0} 0.446} $\pm$ {\color[RGB]{0,0,0} 0.010} & {\color[RGB]{0,0,0} 0.205} $\pm$ {\color[RGB]{0,0,0} 0.012} \\
0.65 & {\color[RGB]{0,153,0} 25.9\%} & {\color[RGB]{255,0,0} 0.375} $\pm$ {\color[RGB]{0,0,0} 0.007} & {\color[RGB]{223,0,0} 0.467} $\pm$ {\color[RGB]{0,0,0} 0.010} & {\color[RGB]{147,0,0} 0.194} $\pm$ {\color[RGB]{0,0,0} 0.012} \\
0.70 & {\color[RGB]{0,144,0} 21.3\%} & {\color[RGB]{238,0,0} 0.406} $\pm$ {\color[RGB]{0,0,0} 0.007} & {\color[RGB]{188,0,0} 0.478} $\pm$ {\color[RGB]{0,0,0} 0.010} & {\color[RGB]{0,0,0} 0.201} $\pm$ {\color[RGB]{0,0,0} 0.012} \\
0.75 & {\color[RGB]{0,135,0} 17.1\%} & {\color[RGB]{185,0,0} 0.423} $\pm$ {\color[RGB]{0,0,0} 0.007} & {\color[RGB]{152,0,0} 0.490} $\pm$ {\color[RGB]{0,0,0} 0.010} & {\color[RGB]{0,0,0} 0.205} $\pm$ {\color[RGB]{0,0,0} 0.012} \\
0.80 & {\color[RGB]{0,128,0} 13.6\%} & {\color[RGB]{152,0,0} 0.434} $\pm$ {\color[RGB]{0,0,0} 0.007} & {\color[RGB]{0,0,0} 0.499} $\pm$ {\color[RGB]{0,0,0} 0.010} & {\color[RGB]{0,0,0} 0.205} $\pm$ {\color[RGB]{0,0,0} 0.012} \\
0.85 & {\color[RGB]{0,121,0} 10.2\%} & {\color[RGB]{127,0,0} 0.442} $\pm$ {\color[RGB]{0,0,0} 0.007} & {\color[RGB]{0,0,0} 0.505} $\pm$ {\color[RGB]{0,0,0} 0.010} & {\color[RGB]{0,0,0} 0.201} $\pm$ {\color[RGB]{0,0,0} 0.012} \\
0.90 & {\color[RGB]{0,115,0} 7.5\%} & {\color[RGB]{0,0,0} 0.449} $\pm$ {\color[RGB]{0,0,0} 0.007} & {\color[RGB]{0,0,0} 0.506} $\pm$ {\color[RGB]{0,0,0} 0.010} & {\color[RGB]{0,0,0} 0.208} $\pm$ {\color[RGB]{0,0,0} 0.012} \\
0.95 & {\color[RGB]{0,108,0} 4.3\%} & {\color[RGB]{0,0,0} 0.451} $\pm$ {\color[RGB]{0,0,0} 0.007} & {\color[RGB]{0,0,0} 0.507} $\pm$ {\color[RGB]{0,0,0} 0.010} & {\color[RGB]{0,0,0} 0.211} $\pm$ {\color[RGB]{0,0,0} 0.012} \\
1.00 & {\color[RGB]{0,0,0} 0.0\%} & {\color[RGB]{0,0,0} 0.451} $\pm$ {\color[RGB]{0,0,0} 0.007} & {\color[RGB]{0,0,0} 0.507} $\pm$ {\color[RGB]{0,0,0} 0.010} & {\color[RGB]{0,0,0} 0.209} $\pm$ {\color[RGB]{0,0,0} 0.012} \\
\hline
\end{tabular}
\label{tab:pythia410m_part2}
\end{table}
\begin{table}[htbp!]
\footnotesize
\centering
\caption{Results for Pythia-1B (Part 1/2)}
\begin{tabular}{lcccccc}
\hline
$\epsilon$ & Speedup & wsc & winogrande & sciq & piqa & logiqa \\
\hline
0.50 & {\color[RGB]{0,201,0} 49.4\%} & {\color[RGB]{0,0,0} 0.365} $\pm$ {\color[RGB]{0,0,0} 0.047} & {\color[RGB]{0,0,0} 0.537} $\pm$ {\color[RGB]{0,0,0} 0.014} & {\color[RGB]{255,0,0} 0.573} $\pm$ {\color[RGB]{0,0,0} 0.016} & {\color[RGB]{255,0,0} 0.613} $\pm$ {\color[RGB]{0,0,0} 0.011} & {\color[RGB]{0,0,0} 0.214} $\pm$ {\color[RGB]{0,0,0} 0.016} \\
0.60 & {\color[RGB]{0,173,0} 35.5\%} & {\color[RGB]{0,0,0} 0.365} $\pm$ {\color[RGB]{0,0,0} 0.047} & {\color[RGB]{0,0,0} 0.527} $\pm$ {\color[RGB]{0,0,0} 0.014} & {\color[RGB]{255,0,0} 0.800} $\pm$ {\color[RGB]{0,0,0} 0.013} & {\color[RGB]{255,0,0} 0.643} $\pm$ {\color[RGB]{0,0,0} 0.011} & {\color[RGB]{0,180,0} 0.235} $\pm$ {\color[RGB]{0,0,0} 0.017} \\
0.65 & {\color[RGB]{0,161,0} 29.8\%} & {\color[RGB]{0,0,0} 0.365} $\pm$ {\color[RGB]{0,0,0} 0.047} & {\color[RGB]{166,0,0} 0.511} $\pm$ {\color[RGB]{0,0,0} 0.014} & {\color[RGB]{230,0,0} 0.834} $\pm$ {\color[RGB]{0,0,0} 0.012} & {\color[RGB]{245,0,0} 0.655} $\pm$ {\color[RGB]{0,0,0} 0.011} & {\color[RGB]{0,0,0} 0.221} $\pm$ {\color[RGB]{0,0,0} 0.016} \\
0.70 & {\color[RGB]{0,151,0} 24.8\%} & {\color[RGB]{0,0,0} 0.365} $\pm$ {\color[RGB]{0,0,0} 0.047} & {\color[RGB]{0,0,0} 0.527} $\pm$ {\color[RGB]{0,0,0} 0.014} & {\color[RGB]{177,0,0} 0.851} $\pm$ {\color[RGB]{0,0,0} 0.011} & {\color[RGB]{233,0,0} 0.659} $\pm$ {\color[RGB]{0,0,0} 0.011} & {\color[RGB]{0,0,0} 0.220} $\pm$ {\color[RGB]{0,0,0} 0.016} \\
0.75 & {\color[RGB]{0,141,0} 20.2\%} & {\color[RGB]{0,0,0} 0.365} $\pm$ {\color[RGB]{0,0,0} 0.047} & {\color[RGB]{144,0,0} 0.518} $\pm$ {\color[RGB]{0,0,0} 0.014} & {\color[RGB]{152,0,0} 0.859} $\pm$ {\color[RGB]{0,0,0} 0.011} & {\color[RGB]{165,0,0} 0.681} $\pm$ {\color[RGB]{0,0,0} 0.011} & {\color[RGB]{0,0,0} 0.223} $\pm$ {\color[RGB]{0,0,0} 0.016} \\
0.80 & {\color[RGB]{0,133,0} 16.3\%} & {\color[RGB]{0,0,0} 0.365} $\pm$ {\color[RGB]{0,0,0} 0.047} & {\color[RGB]{0,0,0} 0.534} $\pm$ {\color[RGB]{0,0,0} 0.014} & {\color[RGB]{152,0,0} 0.859} $\pm$ {\color[RGB]{0,0,0} 0.011} & {\color[RGB]{164,0,0} 0.681} $\pm$ {\color[RGB]{0,0,0} 0.011} & {\color[RGB]{171,0,0} 0.186} $\pm$ {\color[RGB]{0,0,0} 0.015} \\
0.85 & {\color[RGB]{0,126,0} 12.7\%} & {\color[RGB]{0,0,0} 0.365} $\pm$ {\color[RGB]{0,0,0} 0.047} & {\color[RGB]{0,0,0} 0.538} $\pm$ {\color[RGB]{0,0,0} 0.014} & {\color[RGB]{0,0,0} 0.867} $\pm$ {\color[RGB]{0,0,0} 0.011} & {\color[RGB]{133,0,0} 0.691} $\pm$ {\color[RGB]{0,0,0} 0.011} & {\color[RGB]{0,0,0} 0.194} $\pm$ {\color[RGB]{0,0,0} 0.015} \\
0.90 & {\color[RGB]{0,118,0} 8.9\%} & {\color[RGB]{0,0,0} 0.365} $\pm$ {\color[RGB]{0,0,0} 0.047} & {\color[RGB]{0,0,0} 0.534} $\pm$ {\color[RGB]{0,0,0} 0.014} & {\color[RGB]{0,0,0} 0.871} $\pm$ {\color[RGB]{0,0,0} 0.011} & {\color[RGB]{0,0,0} 0.694} $\pm$ {\color[RGB]{0,0,0} 0.011} & {\color[RGB]{0,0,0} 0.210} $\pm$ {\color[RGB]{0,0,0} 0.016} \\
0.95 & {\color[RGB]{0,111,0} 5.4\%} & {\color[RGB]{0,0,0} 0.365} $\pm$ {\color[RGB]{0,0,0} 0.047} & {\color[RGB]{0,0,0} 0.534} $\pm$ {\color[RGB]{0,0,0} 0.014} & {\color[RGB]{0,0,0} 0.874} $\pm$ {\color[RGB]{0,0,0} 0.010} & {\color[RGB]{0,0,0} 0.699} $\pm$ {\color[RGB]{0,0,0} 0.011} & {\color[RGB]{0,0,0} 0.206} $\pm$ {\color[RGB]{0,0,0} 0.016} \\
1.00 & {\color[RGB]{0,0,0} 0.0\%} & {\color[RGB]{0,0,0} 0.365} $\pm$ {\color[RGB]{0,0,0} 0.047} & {\color[RGB]{0,0,0} 0.532} $\pm$ {\color[RGB]{0,0,0} 0.014} & {\color[RGB]{0,0,0} 0.876} $\pm$ {\color[RGB]{0,0,0} 0.010} & {\color[RGB]{0,0,0} 0.702} $\pm$ {\color[RGB]{0,0,0} 0.011} & {\color[RGB]{0,0,0} 0.209} $\pm$ {\color[RGB]{0,0,0} 0.016} \\
\hline
\end{tabular}
\label{tab:pythia1b_part1}
\end{table}
\begin{table}[htbp!]
\centering
\caption{Results for Pythia-1B (Part 2/2)}
\begin{tabular}{lcccc}
\hline
$\epsilon$ & Speedup & lambada\_openai & arc\_easy & arc\_challenge \\
\hline
0.50 & {\color[RGB]{0,201,0} 49.4\%} & {\color[RGB]{255,0,0} 0.276} $\pm$ {\color[RGB]{0,0,0} 0.006} & {\color[RGB]{255,0,0} 0.414} $\pm$ {\color[RGB]{0,0,0} 0.010} & {\color[RGB]{226,0,0} 0.206} $\pm$ {\color[RGB]{0,0,0} 0.012} \\
0.60 & {\color[RGB]{0,173,0} 35.5\%} & {\color[RGB]{255,0,0} 0.417} $\pm$ {\color[RGB]{0,0,0} 0.007} & {\color[RGB]{255,0,0} 0.503} $\pm$ {\color[RGB]{0,0,0} 0.010} & {\color[RGB]{237,0,0} 0.202} $\pm$ {\color[RGB]{0,0,0} 0.012} \\
0.65 & {\color[RGB]{0,161,0} 29.8\%} & {\color[RGB]{255,0,0} 0.465} $\pm$ {\color[RGB]{0,0,0} 0.007} & {\color[RGB]{255,0,0} 0.522} $\pm$ {\color[RGB]{0,0,0} 0.010} & {\color[RGB]{168,0,0} 0.224} $\pm$ {\color[RGB]{0,0,0} 0.012} \\
0.70 & {\color[RGB]{0,151,0} 24.8\%} & {\color[RGB]{255,0,0} 0.508} $\pm$ {\color[RGB]{0,0,0} 0.007} & {\color[RGB]{220,0,0} 0.547} $\pm$ {\color[RGB]{0,0,0} 0.010} & {\color[RGB]{152,0,0} 0.230} $\pm$ {\color[RGB]{0,0,0} 0.012} \\
0.75 & {\color[RGB]{0,141,0} 20.2\%} & {\color[RGB]{206,0,0} 0.541} $\pm$ {\color[RGB]{0,0,0} 0.007} & {\color[RGB]{182,0,0} 0.559} $\pm$ {\color[RGB]{0,0,0} 0.010} & {\color[RGB]{150,0,0} 0.230} $\pm$ {\color[RGB]{0,0,0} 0.012} \\
0.80 & {\color[RGB]{0,133,0} 16.3\%} & {\color[RGB]{153,0,0} 0.558} $\pm$ {\color[RGB]{0,0,0} 0.007} & {\color[RGB]{153,0,0} 0.569} $\pm$ {\color[RGB]{0,0,0} 0.010} & {\color[RGB]{0,0,0} 0.242} $\pm$ {\color[RGB]{0,0,0} 0.013} \\
0.85 & {\color[RGB]{0,126,0} 12.7\%} & {\color[RGB]{125,0,0} 0.567} $\pm$ {\color[RGB]{0,0,0} 0.007} & {\color[RGB]{136,0,0} 0.574} $\pm$ {\color[RGB]{0,0,0} 0.010} & {\color[RGB]{0,0,0} 0.245} $\pm$ {\color[RGB]{0,0,0} 0.013} \\
0.90 & {\color[RGB]{0,118,0} 8.9\%} & {\color[RGB]{0,0,0} 0.574} $\pm$ {\color[RGB]{0,0,0} 0.007} & {\color[RGB]{0,0,0} 0.584} $\pm$ {\color[RGB]{0,0,0} 0.010} & {\color[RGB]{0,0,0} 0.235} $\pm$ {\color[RGB]{0,0,0} 0.012} \\
0.95 & {\color[RGB]{0,111,0} 5.4\%} & {\color[RGB]{0,0,0} 0.576} $\pm$ {\color[RGB]{0,0,0} 0.007} & {\color[RGB]{0,0,0} 0.585} $\pm$ {\color[RGB]{0,0,0} 0.010} & {\color[RGB]{0,0,0} 0.244} $\pm$ {\color[RGB]{0,0,0} 0.013} \\
1.00 & {\color[RGB]{0,0,0} 0.0\%} & {\color[RGB]{0,0,0} 0.575} $\pm$ {\color[RGB]{0,0,0} 0.007} & {\color[RGB]{0,0,0} 0.586} $\pm$ {\color[RGB]{0,0,0} 0.010} & {\color[RGB]{0,0,0} 0.247} $\pm$ {\color[RGB]{0,0,0} 0.013} \\
\hline
\end{tabular}
\label{tab:pythia1b_part2}
\end{table}
\begin{table}[htbp!]
\footnotesize
\centering
\caption{Results for Pythia-1.4B (Part 1/2)}
\begin{tabular}{lcccccc}
\hline
$\epsilon$ & Speedup & wsc & winogrande & sciq & piqa & logiqa \\
\hline
0.50 & {\color[RGB]{0,213,0} 55.0\%} & {\color[RGB]{255,0,0} 0.375} $\pm$ {\color[RGB]{0,0,0} 0.048} & {\color[RGB]{148,0,0} 0.545} $\pm$ {\color[RGB]{0,0,0} 0.014} & {\color[RGB]{255,0,0} 0.502} $\pm$ {\color[RGB]{0,0,0} 0.016} & {\color[RGB]{255,0,0} 0.629} $\pm$ {\color[RGB]{0,0,0} 0.011} & {\color[RGB]{0,0,0} 0.207} $\pm$ {\color[RGB]{0,0,0} 0.016} \\
0.55 & {\color[RGB]{0,198,0} 47.7\%} & {\color[RGB]{255,0,0} 0.404} $\pm$ {\color[RGB]{0,0,0} 0.048} & {\color[RGB]{0,0,0} 0.561} $\pm$ {\color[RGB]{0,0,0} 0.014} & {\color[RGB]{255,0,0} 0.598} $\pm$ {\color[RGB]{0,0,0} 0.016} & {\color[RGB]{255,0,0} 0.628} $\pm$ {\color[RGB]{0,0,0} 0.011} & {\color[RGB]{0,0,0} 0.217} $\pm$ {\color[RGB]{0,0,0} 0.016} \\
0.60 & {\color[RGB]{0,180,0} 39.0\%} & {\color[RGB]{0,0,0} 0.452} $\pm$ {\color[RGB]{0,0,0} 0.049} & {\color[RGB]{0,0,0} 0.556} $\pm$ {\color[RGB]{0,0,0} 0.014} & {\color[RGB]{255,0,0} 0.732} $\pm$ {\color[RGB]{0,0,0} 0.014} & {\color[RGB]{255,0,0} 0.651} $\pm$ {\color[RGB]{0,0,0} 0.011} & {\color[RGB]{0,0,0} 0.217} $\pm$ {\color[RGB]{0,0,0} 0.016} \\
0.65 & {\color[RGB]{0,166,0} 32.4\%} & {\color[RGB]{0,0,0} 0.471} $\pm$ {\color[RGB]{0,0,0} 0.049} & {\color[RGB]{0,0,0} 0.565} $\pm$ {\color[RGB]{0,0,0} 0.014} & {\color[RGB]{255,0,0} 0.798} $\pm$ {\color[RGB]{0,0,0} 0.013} & {\color[RGB]{255,0,0} 0.668} $\pm$ {\color[RGB]{0,0,0} 0.011} & {\color[RGB]{0,0,0} 0.235} $\pm$ {\color[RGB]{0,0,0} 0.017} \\
0.70 & {\color[RGB]{0,156,0} 27.2\%} & {\color[RGB]{0,0,0} 0.471} $\pm$ {\color[RGB]{0,0,0} 0.049} & {\color[RGB]{0,0,0} 0.567} $\pm$ {\color[RGB]{0,0,0} 0.014} & {\color[RGB]{220,0,0} 0.828} $\pm$ {\color[RGB]{0,0,0} 0.012} & {\color[RGB]{243,0,0} 0.675} $\pm$ {\color[RGB]{0,0,0} 0.011} & {\color[RGB]{0,0,0} 0.221} $\pm$ {\color[RGB]{0,0,0} 0.016} \\
0.75 & {\color[RGB]{0,146,0} 22.5\%} & {\color[RGB]{0,0,0} 0.471} $\pm$ {\color[RGB]{0,0,0} 0.049} & {\color[RGB]{0,0,0} 0.557} $\pm$ {\color[RGB]{0,0,0} 0.014} & {\color[RGB]{193,0,0} 0.837} $\pm$ {\color[RGB]{0,0,0} 0.012} & {\color[RGB]{148,0,0} 0.706} $\pm$ {\color[RGB]{0,0,0} 0.011} & {\color[RGB]{0,0,0} 0.233} $\pm$ {\color[RGB]{0,0,0} 0.017} \\
0.80 & {\color[RGB]{0,136,0} 17.8\%} & {\color[RGB]{0,0,0} 0.471} $\pm$ {\color[RGB]{0,0,0} 0.049} & {\color[RGB]{0,0,0} 0.550} $\pm$ {\color[RGB]{0,0,0} 0.014} & {\color[RGB]{162,0,0} 0.847} $\pm$ {\color[RGB]{0,0,0} 0.011} & {\color[RGB]{142,0,0} 0.708} $\pm$ {\color[RGB]{0,0,0} 0.011} & {\color[RGB]{0,0,0} 0.221} $\pm$ {\color[RGB]{0,0,0} 0.016} \\
0.85 & {\color[RGB]{0,128,0} 14.0\%} & {\color[RGB]{0,0,0} 0.471} $\pm$ {\color[RGB]{0,0,0} 0.049} & {\color[RGB]{0,0,0} 0.559} $\pm$ {\color[RGB]{0,0,0} 0.014} & {\color[RGB]{146,0,0} 0.852} $\pm$ {\color[RGB]{0,0,0} 0.011} & {\color[RGB]{145,0,0} 0.707} $\pm$ {\color[RGB]{0,0,0} 0.011} & {\color[RGB]{0,0,0} 0.210} $\pm$ {\color[RGB]{0,0,0} 0.016} \\
0.90 & {\color[RGB]{0,121,0} 10.2\%} & {\color[RGB]{0,0,0} 0.471} $\pm$ {\color[RGB]{0,0,0} 0.049} & {\color[RGB]{0,0,0} 0.557} $\pm$ {\color[RGB]{0,0,0} 0.014} & {\color[RGB]{0,0,0} 0.861} $\pm$ {\color[RGB]{0,0,0} 0.011} & {\color[RGB]{0,0,0} 0.714} $\pm$ {\color[RGB]{0,0,0} 0.011} & {\color[RGB]{152,0,0} 0.204} $\pm$ {\color[RGB]{0,0,0} 0.016} \\
0.95 & {\color[RGB]{0,112,0} 5.9\%} & {\color[RGB]{0,0,0} 0.471} $\pm$ {\color[RGB]{0,0,0} 0.049} & {\color[RGB]{0,0,0} 0.558} $\pm$ {\color[RGB]{0,0,0} 0.014} & {\color[RGB]{0,0,0} 0.866} $\pm$ {\color[RGB]{0,0,0} 0.011} & {\color[RGB]{0,0,0} 0.719} $\pm$ {\color[RGB]{0,0,0} 0.010} & {\color[RGB]{0,0,0} 0.215} $\pm$ {\color[RGB]{0,0,0} 0.016} \\
1.00 & {\color[RGB]{0,0,0} 0.0\%} & {\color[RGB]{0,0,0} 0.471} $\pm$ {\color[RGB]{0,0,0} 0.049} & {\color[RGB]{0,0,0} 0.560} $\pm$ {\color[RGB]{0,0,0} 0.014} & {\color[RGB]{0,0,0} 0.867} $\pm$ {\color[RGB]{0,0,0} 0.011} & {\color[RGB]{0,0,0} 0.721} $\pm$ {\color[RGB]{0,0,0} 0.010} & {\color[RGB]{0,0,0} 0.221} $\pm$ {\color[RGB]{0,0,0} 0.016} \\
\hline
\end{tabular}
\label{tab:pythia1.4b_part1}
\end{table}
\begin{table}[htbp!]
\centering
\caption{Results for Pythia-1.4B (Part 2/2)}
\begin{tabular}{lcccc}
\hline
$\epsilon$ & Speedup & lambada\_openai & arc\_easy & arc\_challenge \\
\hline
0.50 & {\color[RGB]{0,213,0} 55.0\%} & {\color[RGB]{255,0,0} 0.246} $\pm$ {\color[RGB]{0,0,0} 0.006} & {\color[RGB]{255,0,0} 0.399} $\pm$ {\color[RGB]{0,0,0} 0.010} & {\color[RGB]{255,0,0} 0.201} $\pm$ {\color[RGB]{0,0,0} 0.012} \\
0.55 & {\color[RGB]{0,198,0} 47.7\%} & {\color[RGB]{255,0,0} 0.322} $\pm$ {\color[RGB]{0,0,0} 0.007} & {\color[RGB]{255,0,0} 0.447} $\pm$ {\color[RGB]{0,0,0} 0.010} & {\color[RGB]{255,0,0} 0.207} $\pm$ {\color[RGB]{0,0,0} 0.012} \\
0.60 & {\color[RGB]{0,180,0} 39.0\%} & {\color[RGB]{255,0,0} 0.395} $\pm$ {\color[RGB]{0,0,0} 0.007} & {\color[RGB]{255,0,0} 0.508} $\pm$ {\color[RGB]{0,0,0} 0.010} & {\color[RGB]{216,0,0} 0.227} $\pm$ {\color[RGB]{0,0,0} 0.012} \\
0.65 & {\color[RGB]{0,166,0} 32.4\%} & {\color[RGB]{255,0,0} 0.458} $\pm$ {\color[RGB]{0,0,0} 0.007} & {\color[RGB]{255,0,0} 0.542} $\pm$ {\color[RGB]{0,0,0} 0.010} & {\color[RGB]{219,0,0} 0.226} $\pm$ {\color[RGB]{0,0,0} 0.012} \\
0.70 & {\color[RGB]{0,156,0} 27.2\%} & {\color[RGB]{255,0,0} 0.507} $\pm$ {\color[RGB]{0,0,0} 0.007} & {\color[RGB]{255,0,0} 0.558} $\pm$ {\color[RGB]{0,0,0} 0.010} & {\color[RGB]{192,0,0} 0.235} $\pm$ {\color[RGB]{0,0,0} 0.012} \\
0.75 & {\color[RGB]{0,146,0} 22.5\%} & {\color[RGB]{246,0,0} 0.542} $\pm$ {\color[RGB]{0,0,0} 0.007} & {\color[RGB]{246,0,0} 0.570} $\pm$ {\color[RGB]{0,0,0} 0.010} & {\color[RGB]{0,0,0} 0.253} $\pm$ {\color[RGB]{0,0,0} 0.013} \\
0.80 & {\color[RGB]{0,136,0} 17.8\%} & {\color[RGB]{173,0,0} 0.565} $\pm$ {\color[RGB]{0,0,0} 0.007} & {\color[RGB]{184,0,0} 0.590} $\pm$ {\color[RGB]{0,0,0} 0.010} & {\color[RGB]{0,0,0} 0.253} $\pm$ {\color[RGB]{0,0,0} 0.013} \\
0.85 & {\color[RGB]{0,128,0} 14.0\%} & {\color[RGB]{136,0,0} 0.577} $\pm$ {\color[RGB]{0,0,0} 0.007} & {\color[RGB]{140,0,0} 0.604} $\pm$ {\color[RGB]{0,0,0} 0.010} & {\color[RGB]{0,0,0} 0.262} $\pm$ {\color[RGB]{0,0,0} 0.013} \\
0.90 & {\color[RGB]{0,121,0} 10.2\%} & {\color[RGB]{0,0,0} 0.585} $\pm$ {\color[RGB]{0,0,0} 0.007} & {\color[RGB]{0,0,0} 0.613} $\pm$ {\color[RGB]{0,0,0} 0.010} & {\color[RGB]{0,0,0} 0.263} $\pm$ {\color[RGB]{0,0,0} 0.013} \\
0.95 & {\color[RGB]{0,112,0} 5.9\%} & {\color[RGB]{0,0,0} 0.588} $\pm$ {\color[RGB]{0,0,0} 0.007} & {\color[RGB]{0,0,0} 0.615} $\pm$ {\color[RGB]{0,0,0} 0.010} & {\color[RGB]{0,0,0} 0.261} $\pm$ {\color[RGB]{0,0,0} 0.013} \\
1.00 & {\color[RGB]{0,0,0} 0.0\%} & {\color[RGB]{0,0,0} 0.589} $\pm$ {\color[RGB]{0,0,0} 0.007} & {\color[RGB]{0,0,0} 0.617} $\pm$ {\color[RGB]{0,0,0} 0.010} & {\color[RGB]{0,0,0} 0.265} $\pm$ {\color[RGB]{0,0,0} 0.013} \\
\hline
\end{tabular}
\label{tab:pythia1.4b_part1_appendix}
\end{table}
\begin{table}[htbp!]
\footnotesize
\centering
\caption{Results for Pythia-2.8B (Part 1/2)}
\begin{tabular}{lcccccc}
\hline
$\epsilon$ & Speedup & wsc & winogrande & sciq & piqa & logiqa \\
\hline
0.50 & {\color[RGB]{0,221,0} 58.6\%} & {\color[RGB]{0,0,0} 0.394} $\pm$ {\color[RGB]{0,0,0} 0.048} & {\color[RGB]{192,0,0} 0.541} $\pm$ {\color[RGB]{0,0,0} 0.014} & {\color[RGB]{255,0,0} 0.436} $\pm$ {\color[RGB]{0,0,0} 0.016} & {\color[RGB]{255,0,0} 0.615} $\pm$ {\color[RGB]{0,0,0} 0.011} & {\color[RGB]{0,166,0} 0.233} $\pm$ {\color[RGB]{0,0,0} 0.017} \\
0.55 & {\color[RGB]{0,205,0} 51.2\%} & {\color[RGB]{0,255,0} 0.462} $\pm$ {\color[RGB]{0,0,0} 0.049} & {\color[RGB]{0,0,0} 0.564} $\pm$ {\color[RGB]{0,0,0} 0.014} & {\color[RGB]{255,0,0} 0.554} $\pm$ {\color[RGB]{0,0,0} 0.016} & {\color[RGB]{255,0,0} 0.635} $\pm$ {\color[RGB]{0,0,0} 0.011} & {\color[RGB]{0,214,0} 0.249} $\pm$ {\color[RGB]{0,0,0} 0.017} \\
0.60 & {\color[RGB]{0,190,0} 43.7\%} & {\color[RGB]{0,0,0} 0.423} $\pm$ {\color[RGB]{0,0,0} 0.049} & {\color[RGB]{0,0,0} 0.565} $\pm$ {\color[RGB]{0,0,0} 0.014} & {\color[RGB]{255,0,0} 0.688} $\pm$ {\color[RGB]{0,0,0} 0.015} & {\color[RGB]{255,0,0} 0.643} $\pm$ {\color[RGB]{0,0,0} 0.011} & {\color[RGB]{0,176,0} 0.237} $\pm$ {\color[RGB]{0,0,0} 0.017} \\
0.65 & {\color[RGB]{0,174,0} 36.2\%} & {\color[RGB]{0,0,0} 0.404} $\pm$ {\color[RGB]{0,0,0} 0.048} & {\color[RGB]{0,0,0} 0.565} $\pm$ {\color[RGB]{0,0,0} 0.014} & {\color[RGB]{255,0,0} 0.767} $\pm$ {\color[RGB]{0,0,0} 0.013} & {\color[RGB]{255,0,0} 0.668} $\pm$ {\color[RGB]{0,0,0} 0.011} & {\color[RGB]{0,0,0} 0.226} $\pm$ {\color[RGB]{0,0,0} 0.016} \\
0.70 & {\color[RGB]{0,162,0} 30.3\%} & {\color[RGB]{0,0,0} 0.385} $\pm$ {\color[RGB]{0,0,0} 0.048} & {\color[RGB]{0,144,0} 0.586} $\pm$ {\color[RGB]{0,0,0} 0.014} & {\color[RGB]{255,0,0} 0.813} $\pm$ {\color[RGB]{0,0,0} 0.012} & {\color[RGB]{255,0,0} 0.681} $\pm$ {\color[RGB]{0,0,0} 0.011} & {\color[RGB]{0,152,0} 0.229} $\pm$ {\color[RGB]{0,0,0} 0.016} \\
0.75 & {\color[RGB]{0,151,0} 25.0\%} & {\color[RGB]{0,0,0} 0.385} $\pm$ {\color[RGB]{0,0,0} 0.048} & {\color[RGB]{0,0,0} 0.579} $\pm$ {\color[RGB]{0,0,0} 0.014} & {\color[RGB]{233,0,0} 0.839} $\pm$ {\color[RGB]{0,0,0} 0.012} & {\color[RGB]{255,0,0} 0.690} $\pm$ {\color[RGB]{0,0,0} 0.011} & {\color[RGB]{0,0,0} 0.217} $\pm$ {\color[RGB]{0,0,0} 0.016} \\
0.80 & {\color[RGB]{0,141,0} 20.2\%} & {\color[RGB]{0,0,0} 0.385} $\pm$ {\color[RGB]{0,0,0} 0.048} & {\color[RGB]{0,0,0} 0.574} $\pm$ {\color[RGB]{0,0,0} 0.014} & {\color[RGB]{189,0,0} 0.853} $\pm$ {\color[RGB]{0,0,0} 0.011} & {\color[RGB]{207,0,0} 0.707} $\pm$ {\color[RGB]{0,0,0} 0.011} & {\color[RGB]{152,0,0} 0.195} $\pm$ {\color[RGB]{0,0,0} 0.016} \\
0.85 & {\color[RGB]{0,132,0} 15.8\%} & {\color[RGB]{0,0,0} 0.385} $\pm$ {\color[RGB]{0,0,0} 0.048} & {\color[RGB]{0,0,0} 0.579} $\pm$ {\color[RGB]{0,0,0} 0.014} & {\color[RGB]{146,0,0} 0.867} $\pm$ {\color[RGB]{0,0,0} 0.011} & {\color[RGB]{170,0,0} 0.719} $\pm$ {\color[RGB]{0,0,0} 0.010} & {\color[RGB]{152,0,0} 0.195} $\pm$ {\color[RGB]{0,0,0} 0.016} \\
0.90 & {\color[RGB]{0,124,0} 12.0\%} & {\color[RGB]{0,0,0} 0.385} $\pm$ {\color[RGB]{0,0,0} 0.048} & {\color[RGB]{0,0,0} 0.571} $\pm$ {\color[RGB]{0,0,0} 0.014} & {\color[RGB]{146,0,0} 0.867} $\pm$ {\color[RGB]{0,0,0} 0.011} & {\color[RGB]{135,0,0} 0.730} $\pm$ {\color[RGB]{0,0,0} 0.010} & {\color[RGB]{152,0,0} 0.195} $\pm$ {\color[RGB]{0,0,0} 0.016} \\
0.95 & {\color[RGB]{0,115,0} 7.6\%} & {\color[RGB]{0,0,0} 0.385} $\pm$ {\color[RGB]{0,0,0} 0.048} & {\color[RGB]{0,0,0} 0.572} $\pm$ {\color[RGB]{0,0,0} 0.014} & {\color[RGB]{137,0,0} 0.870} $\pm$ {\color[RGB]{0,0,0} 0.011} & {\color[RGB]{0,0,0} 0.733} $\pm$ {\color[RGB]{0,0,0} 0.010} & {\color[RGB]{0,0,0} 0.201} $\pm$ {\color[RGB]{0,0,0} 0.016} \\
1.00 & {\color[RGB]{0,0,0} 0.0\%} & {\color[RGB]{0,0,0} 0.385} $\pm$ {\color[RGB]{0,0,0} 0.048} & {\color[RGB]{0,0,0} 0.571} $\pm$ {\color[RGB]{0,0,0} 0.014} & {\color[RGB]{0,0,0} 0.882} $\pm$ {\color[RGB]{0,0,0} 0.010} & {\color[RGB]{0,0,0} 0.742} $\pm$ {\color[RGB]{0,0,0} 0.010} & {\color[RGB]{0,0,0} 0.212} $\pm$ {\color[RGB]{0,0,0} 0.016} \\
\hline
\end{tabular}
\label{tab:pythia2.8b_part1_appendix}
\end{table}
\begin{table}[htbp!]
\centering
\caption{Results for Pythia-2.8B (Part 2/2)}
\begin{tabular}{lcccc}
\hline
$\epsilon$ & Speedup & lambada\_openai & arc\_easy & arc\_challenge \\
\hline
0.50 & {\color[RGB]{0,221,0} 58.6\%} & {\color[RGB]{255,0,0} 0.249} $\pm$ {\color[RGB]{0,0,0} 0.006} & {\color[RGB]{255,0,0} 0.407} $\pm$ {\color[RGB]{0,0,0} 0.010} & {\color[RGB]{255,0,0} 0.217} $\pm$ {\color[RGB]{0,0,0} 0.012} \\
0.55 & {\color[RGB]{0,205,0} 51.2\%} & {\color[RGB]{255,0,0} 0.340} $\pm$ {\color[RGB]{0,0,0} 0.007} & {\color[RGB]{255,0,0} 0.450} $\pm$ {\color[RGB]{0,0,0} 0.010} & {\color[RGB]{255,0,0} 0.238} $\pm$ {\color[RGB]{0,0,0} 0.012} \\
0.60 & {\color[RGB]{0,190,0} 43.7\%} & {\color[RGB]{255,0,0} 0.416} $\pm$ {\color[RGB]{0,0,0} 0.007} & {\color[RGB]{255,0,0} 0.495} $\pm$ {\color[RGB]{0,0,0} 0.010} & {\color[RGB]{255,0,0} 0.243} $\pm$ {\color[RGB]{0,0,0} 0.013} \\
0.65 & {\color[RGB]{0,174,0} 36.2\%} & {\color[RGB]{255,0,0} 0.482} $\pm$ {\color[RGB]{0,0,0} 0.007} & {\color[RGB]{255,0,0} 0.537} $\pm$ {\color[RGB]{0,0,0} 0.010} & {\color[RGB]{158,0,0} 0.276} $\pm$ {\color[RGB]{0,0,0} 0.013} \\
0.70 & {\color[RGB]{0,162,0} 30.3\%} & {\color[RGB]{255,0,0} 0.533} $\pm$ {\color[RGB]{0,0,0} 0.007} & {\color[RGB]{255,0,0} 0.565} $\pm$ {\color[RGB]{0,0,0} 0.010} & {\color[RGB]{155,0,0} 0.276} $\pm$ {\color[RGB]{0,0,0} 0.013} \\
0.75 & {\color[RGB]{0,151,0} 25.0\%} & {\color[RGB]{255,0,0} 0.570} $\pm$ {\color[RGB]{0,0,0} 0.007} & {\color[RGB]{243,0,0} 0.594} $\pm$ {\color[RGB]{0,0,0} 0.010} & {\color[RGB]{0,0,0} 0.288} $\pm$ {\color[RGB]{0,0,0} 0.013} \\
0.80 & {\color[RGB]{0,141,0} 20.2\%} & {\color[RGB]{225,0,0} 0.596} $\pm$ {\color[RGB]{0,0,0} 0.007} & {\color[RGB]{192,0,0} 0.610} $\pm$ {\color[RGB]{0,0,0} 0.010} & {\color[RGB]{0,0,0} 0.291} $\pm$ {\color[RGB]{0,0,0} 0.013} \\
0.85 & {\color[RGB]{0,132,0} 15.8\%} & {\color[RGB]{157,0,0} 0.618} $\pm$ {\color[RGB]{0,0,0} 0.007} & {\color[RGB]{158,0,0} 0.621} $\pm$ {\color[RGB]{0,0,0} 0.010} & {\color[RGB]{0,0,0} 0.299} $\pm$ {\color[RGB]{0,0,0} 0.013} \\
0.90 & {\color[RGB]{0,124,0} 12.0\%} & {\color[RGB]{0,0,0} 0.630} $\pm$ {\color[RGB]{0,0,0} 0.007} & {\color[RGB]{0,0,0} 0.630} $\pm$ {\color[RGB]{0,0,0} 0.010} & {\color[RGB]{0,0,0} 0.294} $\pm$ {\color[RGB]{0,0,0} 0.013} \\
0.95 & {\color[RGB]{0,115,0} 7.6\%} & {\color[RGB]{0,0,0} 0.635} $\pm$ {\color[RGB]{0,0,0} 0.007} & {\color[RGB]{0,0,0} 0.636} $\pm$ {\color[RGB]{0,0,0} 0.010} & {\color[RGB]{0,0,0} 0.291} $\pm$ {\color[RGB]{0,0,0} 0.013} \\
1.00 & {\color[RGB]{0,0,0} 0.0\%} & {\color[RGB]{0,0,0} 0.637} $\pm$ {\color[RGB]{0,0,0} 0.007} & {\color[RGB]{0,0,0} 0.640} $\pm$ {\color[RGB]{0,0,0} 0.010} & {\color[RGB]{0,0,0} 0.294} $\pm$ {\color[RGB]{0,0,0} 0.013} \\
\hline
\end{tabular}
\label{tab:pythia2.8b_part2_appendix}
\end{table}

\subsection{Post-training analysis}
\label{subsec:posttraining}

\begin{figure}[ht]
    \centering
    \includegraphics[width=0.8\textwidth]{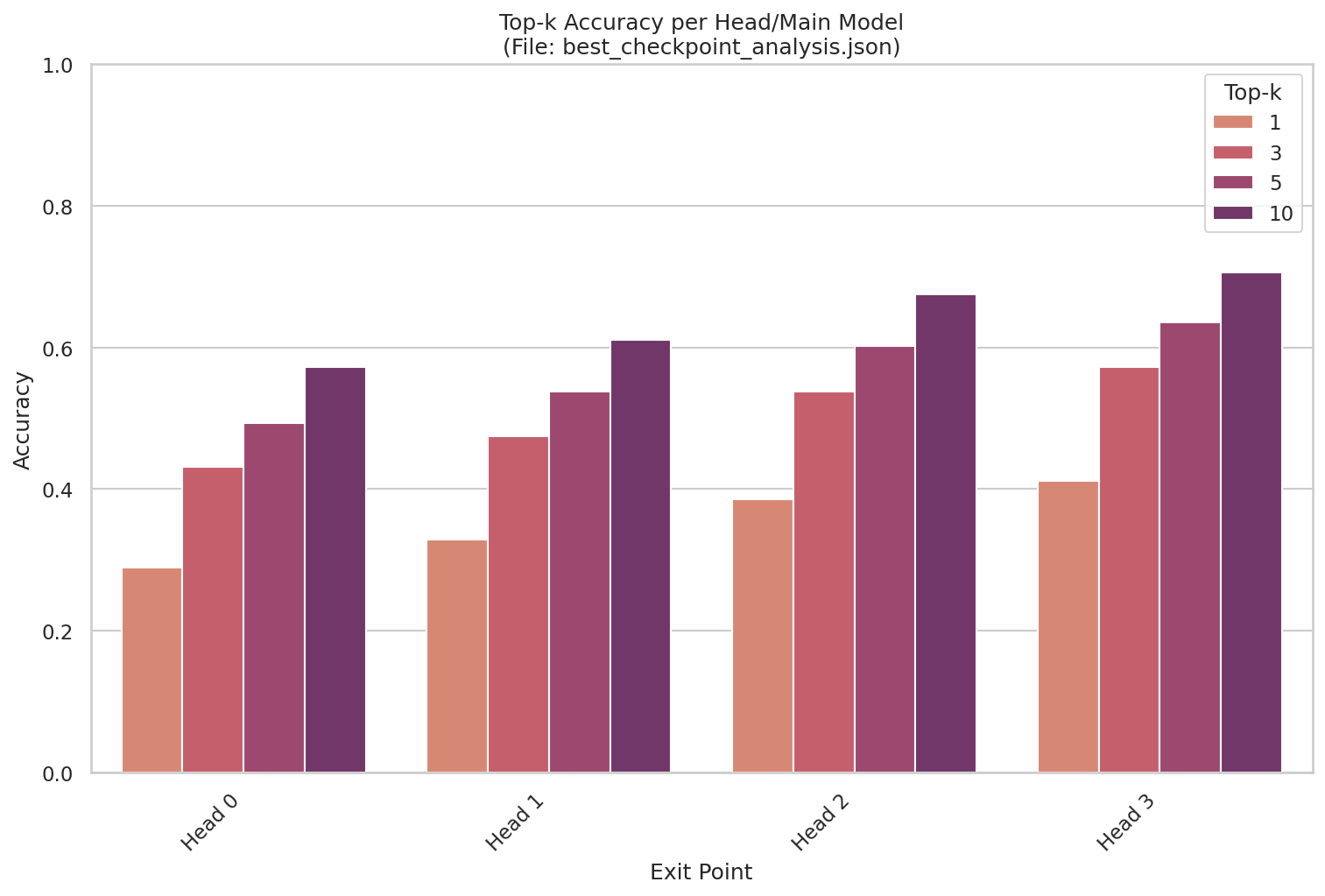}
    \caption{Top-1, top-3, top-5, and top-10 accuracy of each early exit head on Pythia-2.8B after post-training. Intermediate heads achieve competitive accuracy relative to the final head.}
    \label{fig:topk_accuracy_per_head}
\end{figure}

To further analyze the quality of the early exit heads after training, we report the top-1, top-3, top-5, and top-10 accuracy of each head on the Pythia-2.8B model. Figure~\ref{fig:topk_accuracy_per_head} shows that the intermediate heads, post-training, achieve competitive accuracy compared to the final head. This demonstrates that our post-training procedure enables early exit heads to provide strong predictions, making early exits practical and reliable in this setting.

\subsection{Entropy distribution analysis}
\label{subsec:entropyviolin}

\begin{figure}[ht]
    \centering
    \includegraphics[width=0.7\textwidth]{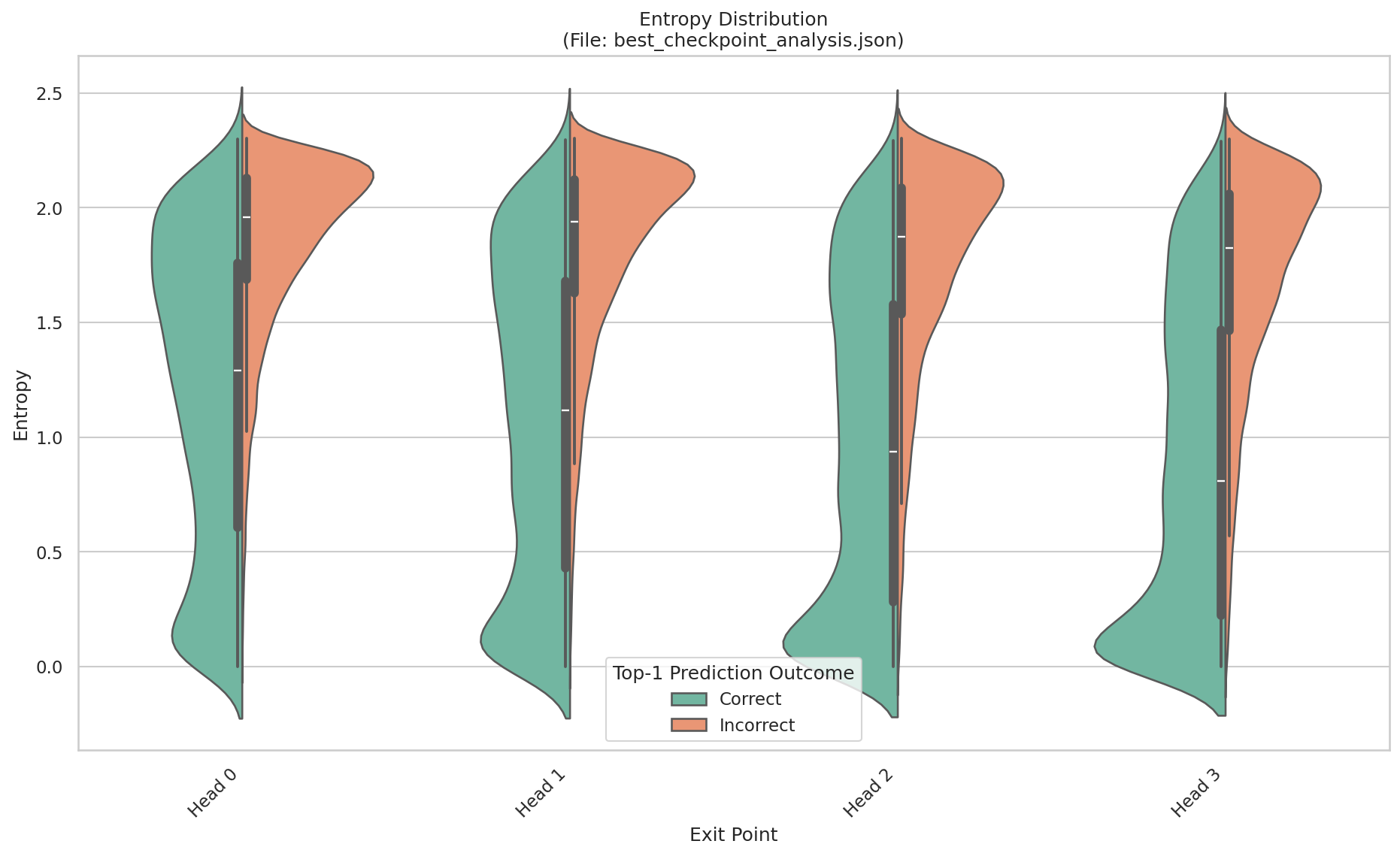}
    \caption{Violin plot of the entropy values for correct and incorrect outputs on the calibration set. The entropy metric provides good separation between right and wrong predictions.}
    \label{fig:entropy_distribution_violin}
\end{figure}

To further illustrate the discriminative power of the entropy metric, Figure~\ref{fig:entropy_distribution_violin} presents a violin plot of the entropy values computed on the calibration set. This plot shows the distribution of entropy for correct and incorrect outputs, demonstrating that the entropy metric can effectively separate right and wrong predictions in our model.

\subsection{Per-token accuracy analysis}
\label{subsec:pertokenaccuracy}

\begin{figure}[ht]
    \centering
    \includegraphics[width=0.6\textwidth]{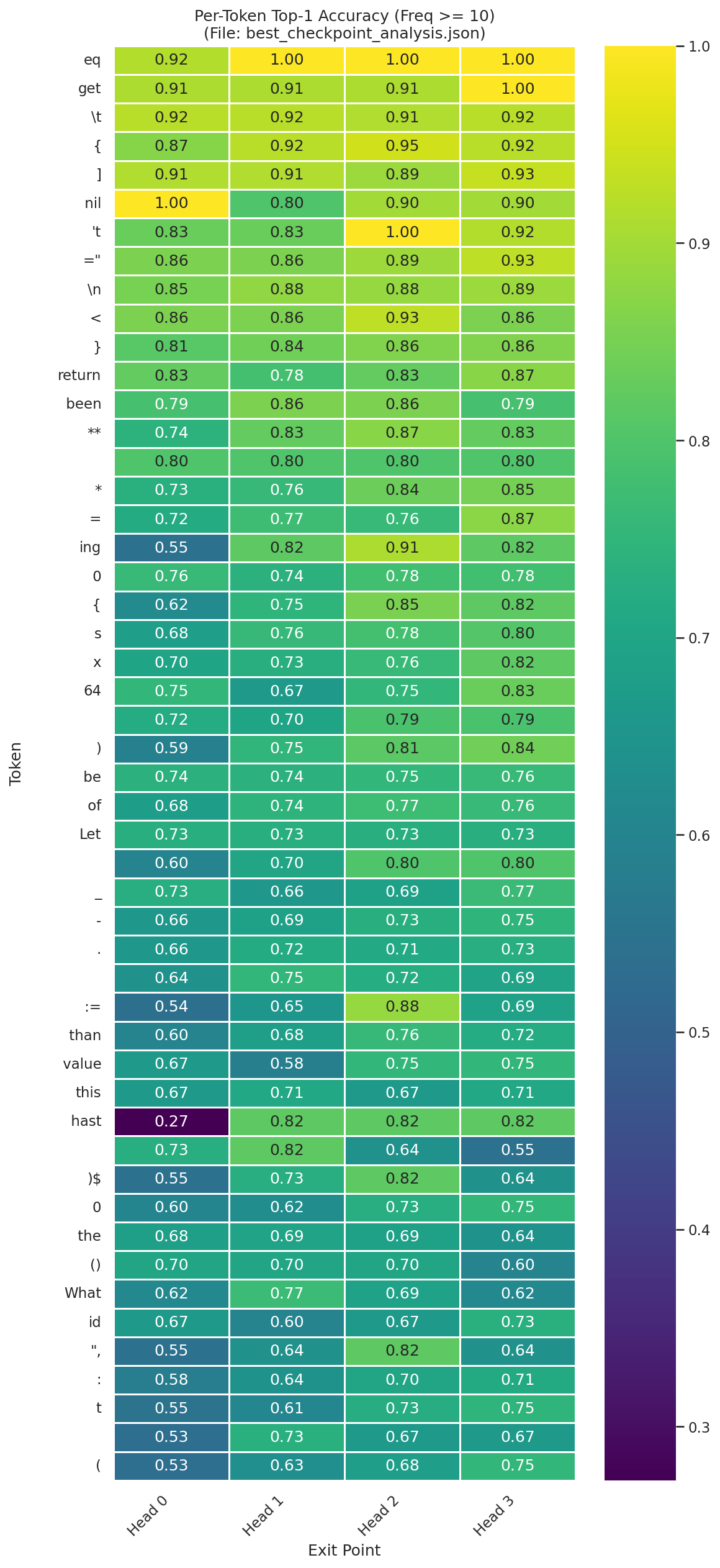}
    \caption{Heatmap of top-1 accuracy per token for each early exit head. Tokens with high accuracy at intermediate heads are typically common or easy tokens, supporting the use of early exit for efficient inference.}
    \label{fig:per_token_top1_accuracy_heatmap}
\end{figure}

To further support the idea that early exit heads are particularly effective on ``easy'' tokens, Figure~\ref{fig:per_token_top1_accuracy_heatmap} presents a heatmap of the top-1 accuracy per token for each head. This plot highlights that certain tokens achieve very high accuracy at intermediate heads, and these tokens often correspond to common or simple tokens in the dataset. This observation confirms that our early exit mechanism is able to leverage the inherent variability in token difficulty: easy tokens can be predicted confidently and accurately by earlier heads, enabling faster inference without sacrificing accuracy on these cases.

\begin{figure}
\centering
\includegraphics[width=\textwidth]{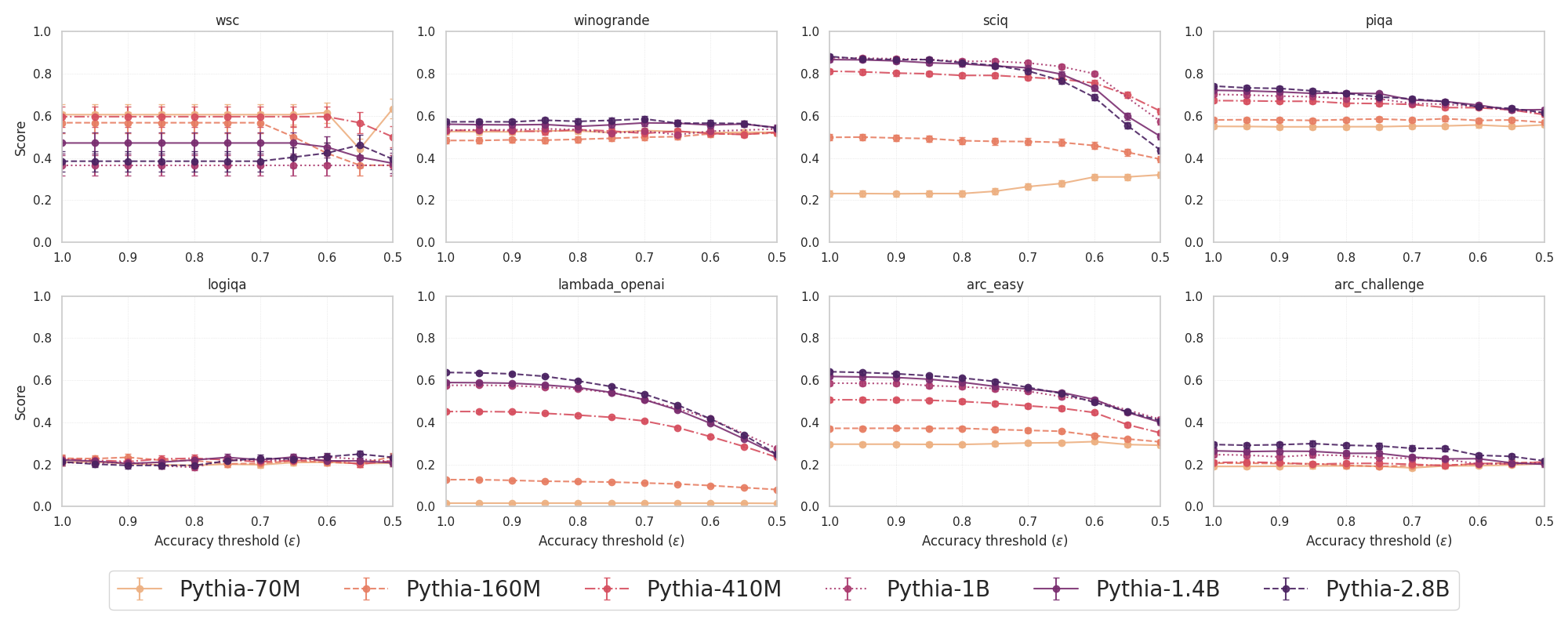}
\caption{Effect of dynamic exiting on benchmark performance.}
\label{benchmarks}
\end{figure}

we show in Figure~\ref{benchmarks} that, when using dynamic exiting, some benchmark tends to have a lower score the more we lower the accuracy threshold. On the other hand, some benchmark does not have this tendency and stay at the same level while getting a speedup from the exiting.

\end{document}